\documentclass[journal]{IEEEtran}
\usepackage{fix-cm}
\usepackage[utf8]{inputenc}
\usepackage{amsmath,amsfonts}
\usepackage{array}
\usepackage[caption=false]{subfig}
\usepackage{textcomp}
\usepackage{stfloats}
\usepackage{url}
\usepackage{verbatim}
\usepackage{graphicx}
\usepackage[T1]{fontenc}
\usepackage{threeparttable}
\usepackage[numbers,sort&compress]{natbib}
\def\BibTeX{{\rm B\kern-.05em{\sc i\kern-.025em b}\kern-.08em
    T\kern-.1667em\lower.7ex\hbox{E}\kern-.125emX}}
\usepackage{balance}
\usepackage{utfsym}
\usepackage{amssymb}
\usepackage{float}
\usepackage{hyperref}
\usepackage{setspace}
\usepackage{float}
\usepackage{bm}
\usepackage{multirow}
\usepackage{colortbl}
\usepackage{xcolor}
\usepackage{arydshln}
\usepackage{algorithm}
\usepackage{algorithmicx}
\usepackage{algpseudocode}
\usepackage{tabularx}
\hypersetup{
	colorlinks,
	linkcolor=blue, 
	citecolor=cyan!50!gray,
	urlcolor=cyan
}
\newcommand{\std}[1]{\textsuperscript{\fontsize{5.8}{6.5}\selectfont $\pm$#1}}
\usepackage{booktabs}  
\usepackage{multirow}   
\usepackage{threeparttable}

\begin{document}
\title{Hierarchical Multimodal Interaction Reasoning for Dyadic Relationship Recognition}

\author{Wang Tang, Fethiye Irmak Dogan, Linbo Qing, Ce Zhu, \textit{Fellow}, \textit{IEEE}, Hatice Gunes, \textit{Senior Member}, \textit{IEEE}
\thanks{Wang Tang is with the School of Software Engineering, Chengdu University of Information Technology, Chengdu 610103, China (e-mail: wangtang@cuit.edu.cn). This work was supported in part by the National Natural Science Foundation of China under Grant 6267073968.}
\thanks{F. Irmak Dogan and Hatice Gunes are with the AFAR Lab, Department of Computer Science and Technology, University of Cambridge, CB3 0FT Cambridge, U.K. (e-mail:\{fid21,hg410\}@cam.ac.uk). They are supported in part by CHANSE and NORFACE through the MICRO project, funded by UKRI/ESRC (grant ref. UKRI572).}
\thanks{Linbo Qing is with the College of Electronics and Information Engineering, Sichuan University, Chengdu 610065, China (e-mail: qing$\_$lb@scu.edu.cn).}
\thanks{Ce Zhu is with the School of Information and Communication Engineering, University of Electronic Science and Technology of China, Chengdu 610051, China (e-mail: eczhu@uestc.edu.cn).}
\thanks{\textbf{Contributions.} Conceptualisation: WT, FID, HG. Methodology: WT, CZ, FID, HG. Software, Validation, Data Curation, Formal Analysis, and Investigation: WT. Writing – Original Draft: WT. Visualisation: WT. Writing – Review \& Editing: WT, LQ, FID, HG. Supervision \& Project Administration: LQ, CZ, HG. Funding Acquisition: LQ \& HG.}
\thanks{\textbf{Data access.} This work undertakes secondary analyses on existing datasets that are cited accordingly in the paper.}
}

\maketitle
\begin{abstract}
Dyadic relationship recognition aims to identify the latent social relationship between two individuals from their multimodal interactions. Existing methods typically project heterogeneous cues into a shared representation and optimise a flat classification objective, overlooking that relationships may rely on different interaction patterns. For instance, distant relationships can often be inferred from direct behavioural correspondence, whereas close relationships are better reflected by coordinated affective, behavioural, and multimodal responses. To address this limitation, we propose HiRe-DRR, a Hierarchical Multimodal Interaction Reasoning framework that organises relationship evidence from three complementary perspectives: feature level, interaction type, and relationship hierarchy. HiRe-DRR decomposes facial, bodily, acoustic, and textual cues into low- and high-level representations and constructs homogeneous and heterogeneous cross-participant interactions. We further introduce the Mixture-of-Interaction Transformer (MoI-Former), which treats diverse interactions as specialised experts and employs relation-aware sparse routing to identify relationship-relevant evidence while preserving contextual interaction information. The proposed Hierarchical Conditional Loss (HC-Loss) exploits the hierarchical nature of social relationships through coarse-to-fine supervision, enabling structured relationship reasoning. Extensive experiments on three dyadic relationship benchmarks demonstrate that HiRe-DRR outperforms recent representative image-, video-, and multimodal methods. Our routing analysis further reveals that distant relationships tend to rely on low-level homogeneous interactions, whereas closer relationships focus on high-level heterogeneous interactions. Our code is publicly available at: \href{https://github.com/tw-repository/HiRe-DRR}{https://github.com/tw-repository/HiRe-DRR}.
\end{abstract}


\section{Introduction}
\IEEEPARstart{H}{uman} dyadic relationship recognition (DRR) aims to infer the latent social relationship between two individuals from their multimodal interactions. Unlike transient actions or emotions, relationships are higher-order social constructs reflected in recurring behavioural patterns, affective exchanges, verbal communication, and social intentions~\cite{01}. For example, interactions between unfamiliar individuals are often characterised by observable behavioural coordination, such as gaze alignment, speaking turns, and physical distance, whereas close relationships may involve richer emotional responses and implicit behavioural adaptation. Understanding such interaction dynamics is fundamental to socially intelligent systems~\cite{03}. In human-robot interaction, relationship-aware agents can adapt their behaviours to their partners' social contexts~\cite{02,05}, while pairwise relationships provide latent structures for interpreting group activities~\cite{08}. DRR also supports affective computing~\cite{09} and social interaction analysis~\cite{10}.

\begin{figure*}[t!]
    \vspace{-0.5em}
    \setlength{\abovecaptionskip}{0cm}
    \setlength{\belowcaptionskip}{0cm}
    \centering
    \includegraphics[width=5.0in]{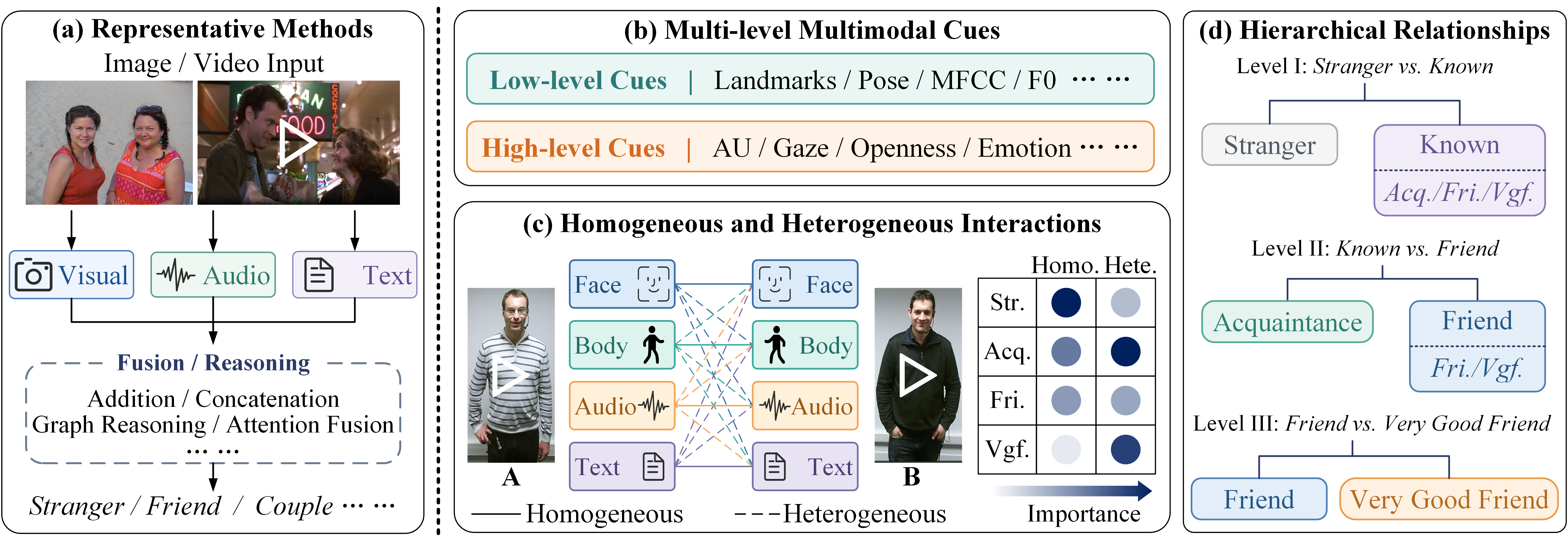}
    \caption{Existing SRR paradigms and three challenges of DRR:
    (a) representative image- and video-based settings;
    (b) multi-level multimodal cues;
    (c) homogeneous and heterogeneous interactions;
    and (d) hierarchical relationship structure. The examples in
    (a) are from PISC~\cite{12} and MovieGraphs~\cite{30}, while those
    in (c) are from NoXi~\cite{31}. \textit{Acq.}, \textit{Fri.}, and
    \textit{Vgf.} denote Acquaintance, Friend, and Very Good Friend,
    respectively.}
    \label{Fig: 1}
    \vspace{-1.0em}
\end{figure*}

Early social relationship recognition (SRR) methods focused mainly on static images, using visual appearance~\cite{12,13}, scene context and person-object interactions~\cite{15,17}. Subsequent video and multimodal approaches incorporated temporal behaviours, vocal cues, and linguistic information~\cite{18,19}, typically projecting them into a shared space for attention- or graph-based fusion~\cite{22,23,24,25,26,29}, as illustrated in Fig.~\ref{Fig: 1}(a). Although these methods improve recognition performance, their holistic fusion generally encodes multimodal associations implicitly, making it difficult to determine which interaction structures characterise a particular relationship.

Recent multimodal foundation models provide transferable representations, broad semantic knowledge, and the ability to produce natural-language explanations~\cite{79,80}. However, improved semantic reasoning does not necessarily reveal how a relationship decision is grounded in participant-specific interaction evidence. Foundation models often rely on globally aggregated multimodal representations, making it difficult to identify which interaction structures contribute to different relationships. For example, distant relationships may be characterised by observable behavioural consistencies, such as facial alignment, speaking patterns, or behavioural synchrony, whereas close relationships may involve subtle emotional responses and richer cross-modal behavioural exchanges between individuals. Therefore, how multimodal interaction evidence at different semantic levels characterises distinct relationships, and how such evidence varies across the relationship hierarchy, remains an open question. We formulate this problem through three aspects: feature level, cross-participant interaction type, and relationship hierarchy.

First, relationship evidence spans multiple semantic levels (Fig.~\ref{Fig: 1}(b))~\cite{32,33}. Low-level cues, including facial landmarks, body pose, acoustic rhythm, and spectral descriptors, capture observable behavioural consistency, such as synchronized expressions, posture alignment, and conversational rhythms~\cite{33}. High-level cues, such as facial action units, gaze, body openness, affective states, and linguistic sentiment, provide richer social semantics by describing emotional responses, engagement, and interpersonal coordination~\cite{33,34}. Directly projecting these cues into a single space may obscure their semantic distinctions and prevent the model from revealing how evidence at different levels contributes to each relationship. (\textit{Challenge} I)

Second, relationships emerge through different cross-participant interaction patterns (Fig.~\ref{Fig: 1}(c)). Homogeneous interactions capture dependencies between the same modalities, such as face-face correspondence~\cite{35}, whereas heterogeneous interactions represent cross-modal responses, such as associations between one participant's speech and the other's facial or bodily reactions~\cite{36}. For example, a stranger may mainly exhibit consistent facial or speech patterns, while close partners may respond through coordinated facial, bodily, and acoustic behaviours. Uniformly aggregating all interactions conceals such relationship-specific patterns. The model must therefore represent these structures separately and adaptively identify those relevant to each relationship. (\textit{Challenge} II)

Third, relationship classes exhibit an inherent hierarchy, rather than independent class boundaries. As shown in Fig.~\ref{Fig: 1}(d), relationship recognition can naturally be decomposed into a sequence of conditional decisions. The model can first determines whether two individuals are acquainted; if they are, it further distinguishes an ordinary \textit{Acquaintance} from \textit{Friend}; once a friendship is determined, it finally determines whether the pair are \textit{Very Good Friends}. However, conventional cross-entropy treats all classes as independent targets and assigns identical penalties to different types of errors, ignoring the semantic proximity among relationship classes. This motivates the third challenge of incorporating relationship hierarchy into the learning objective to supervise hierarchical relationship reasoning. (\textit{Challenge} III)

To address these challenges, we propose HiRe-DRR, a \textbf{Hi}erarchical Multimodal Interaction \textbf{Re}asoning framework for \textbf{D}yadic \textbf{R}elationship \textbf{R}ecognition. HiRe-DRR organises multimodal representations according to their semantic levels and constructs homogeneous and heterogeneous cross-participant interactions at each level. We then propose the \textbf{M}ixture-\textbf{o}f-\textbf{I}nteraction Transformer (MoI-Former), which treats these interaction features as specialised experts and uses top-$K$ sparse routing to select relationship-relevant evidence. Its routing preferences provide an explicit view of how different relationships depend on different feature levels and interaction types. We further introduce a \textbf{H}ierarchical \textbf{C}onditional \textbf{L}oss (HC-Loss), which decomposes supervision into conditional decisions along the relationship hierarchy, connecting interaction representations with coarse-to-fine relationship semantics. Our main contributions are as follows:

\begin{itemize}
    \item We propose HiRe-DRR, a hierarchical multimodal interaction reasoning framework that jointly models feature level, interaction type, and relationship hierarchy for dyadic relationship recognition. (\textit{Addressing Challenge} I, II, III)

    \item We introduce MoI-Former, which treats different interaction cues as specialised relationship experts and employs Top-$K$ sparse routing to adaptively select and aggregate relationship-relevant interaction evidence. \textit{Addressing Challenge} I, II)

    \item We propose HC-Loss, which incorporates relationship hierarchy into the supervisory process via a series of binary conditional decisions, enabling the model to learn hierarchical relationship representations. (\textit{Addressing Challenge} III)

    \item Experiments on NoXi, UDIVA, and Seamless Interaction demonstrate competitive relationship recognition, while ablation and routing analyses reveal distinct feature-level and interaction patterns across relationship classes.
\end{itemize}

\section{RELATED WORK}
\subsection{Static Image-based SRR}
Early image-based SRR inferred interpersonal relationships from visual appearance, facial attributes, pairwise geometry, and scene context~\cite{12,13}. Li \textit{et al.}~\cite{11} combined appearance, facial cues, and interpersonal geometry to represent person pairs, while subsequent methods introduced contextual information for improved relationship discrimination ~\cite{12,13}. More recently, Wu \textit{et al.}~\cite{39} adapted vision-language representations to identify socially informative objects, activities, and contextual cues. These studies extended SRR beyond individual appearance towards contextual semantic reasoning.

Graph-based methods further represent people, objects, and context as structured relational entities~\cite{16,14,22}. Tang \textit{et al.} constructed progressive graphs across individual, pair, and group levels~\cite{42}. Other studies introduced triangular constraints for third-party influence~\cite{15}, multi-level attention across representational scales~\cite{24}, and graph-based knowledge distillation for incremental learning~\cite{43}. Hypergraph Mamba Reasoning~\cite{17} further captured higher-order and long-range dependencies through person-person and person-object hypergraphs.

Large multimodal models have recently connected visual observations with abstract social semantics. SocialGPT~\cite{44} transforms visual scenes into socially grounded descriptions for zero-shot inference, while Wang \textit{et al.}~\cite{45} align visual and linguistic representations for complementary reasoning. GMLLM-SRR~\cite{46} further combines graph-structured visual evidence with multimodal language models through cross-attention. Although these methods provide strong semantic priors and explanatory outputs, their image-based evidence cannot capture recurring behaviours or temporally evolving responses between participants.

\subsection{Spatio-temporal Multimodal SRR}
Video-based SRR extends static inference by modelling evolving social behaviours and their temporal dependencies. Earlier structured approaches represented people, objects, and narrative context over time using pyramid graph convolutions~\cite{18}, dynamic character-relationship graphs~\cite{51}, spatio-temporal message passing~\cite{26}, and multi granularity graph aggregation~\cite{52}. Subsequent multimodal methods incorporated speech, acoustic cues, and linguistic semantics~\cite{19,30} through action-scene attention~\cite{20}, progressive aggregation of frame-level multimodal graphs~\cite{28}, cross-modal pre-training of visual, speech, and textual signals~\cite{27}, and cumulative graph--Transformer reasoning for long-range dependencies~\cite{29}. Recent work has further introduced causal modelling to mitigate culture-specific correlations~\cite{53} and capture dependencies among temporal observations~\cite{54}. Together, these advances have shifted video SRR from conventional sequence encoding towards structured multimodal temporal reasoning; however, most remain character-centric and tailored to movie narratives, with limited modelling of real-world dyadic interactions.

For real-world DRR, recent studies have extended social understanding beyond movie-based scenarios by modelling interpersonal dynamics in natural interactions. AsyReC~\cite{55} captures asymmetric dependencies between participants through multimodal graph reasoning and temporal encoding, highlighting the importance of modelling evolving interaction patterns in dyadic conversations. Meanwhile, multimodal foundation models have introduced new directions for social reasoning by integrating visual, auditory, and linguistic information with large-scale semantic knowledge. Ouyang \textit{et al.}~\cite{80} investigate multimodal social interaction in multi-speaker scenarios and propose attention alignment strategies to strengthen the correspondence between speakers' visual behaviours and utterances. MimeQA~\cite{81} further establishes a benchmark for evaluating nonverbal social intelligence in foundation models, emphasizing the role of embodied gestures and contextual cues in social understanding. FriendBench~\cite{79} directly studies dyadic familiarity inference with multimodal large language models and human comparisons, providing insights into how different modalities contribute to relationship perception.

Collectively, these advances demonstrate that understanding dyadic relationships requires reasoning over rich multimodal social behaviours rather than relying solely on linguistic semantics. However, existing approaches generally focus on improving recognition ability or semantic reasoning, while the organisation of relationship evidence remains largely implicit. It remains unclear which interaction patterns distinguish different relationships and how such evidence is structured across relationship categories. Addressing this issue requires an explicit and interpretable framework that organises multimodal evidence according to its semantic characteristics, interaction structures, and relational hierarchy. In this work, HiRe-DRR addresses this gap by performing hierarchical multimodal interaction reasoning.

\begin{figure*}[t!]
\vspace{-0.5em}
\setlength{\abovecaptionskip}{0.cm}
\setlength{\belowcaptionskip}{-0.cm}
\centering
\includegraphics[width=6.0in]{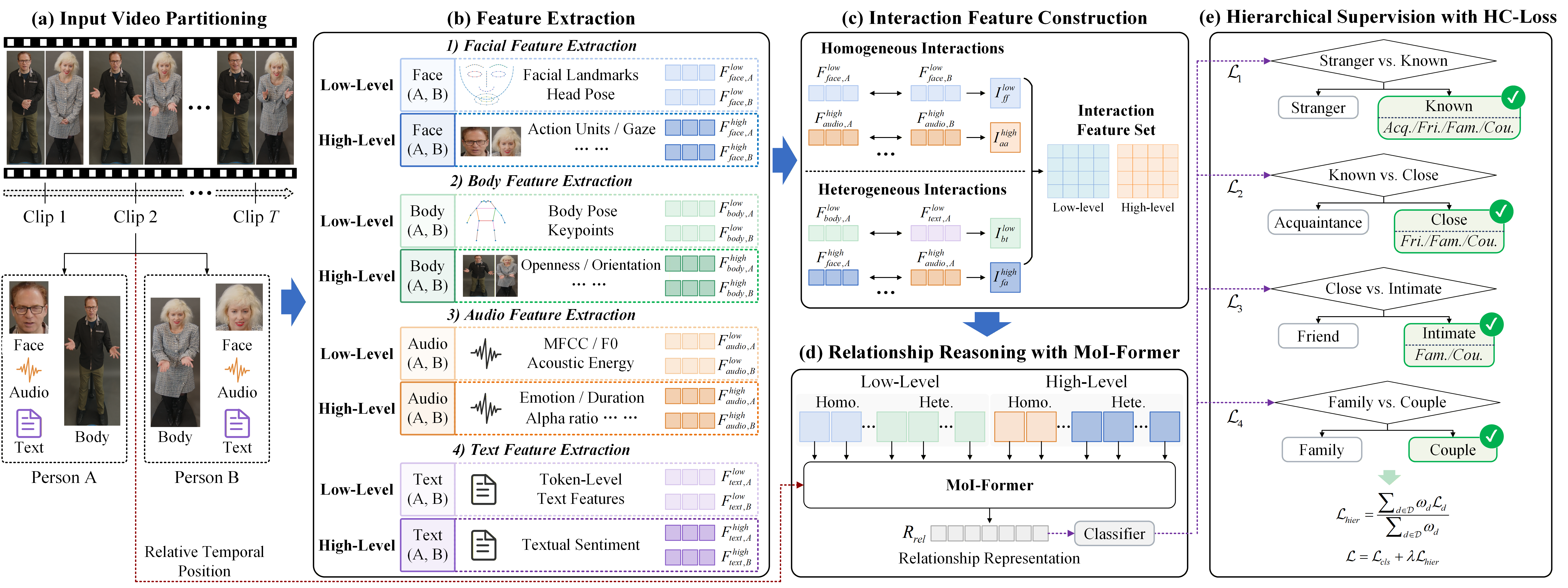}
    \caption{Overview of the proposed HiRe-DRR framework.
    (a) Input video partition.
    (b) Low- and high-level feature extraction.
    (c) Homogeneous and heterogeneous interaction features construction.
    (d) Relationship reasoning with MoI-Former.
    (e) Hierarchical Supervision with HC-Loss.
    The screenshots are taken from the Seamless Interaction dataset~\cite{56}. 
    \textit{Note}: \textit{Acq.}, \textit{Fri.}, \textit{Fam.}, and \textit{Cou.} denote Acquaintance, Friend, Family, and Couple, respectively.}
\label{Fig: 2}
\vspace{-1.0em}
\end{figure*}

\section{Method}
The HiRe-DRR pipeline is shown in Fig.~\ref{Fig: 2}. Each video is divided into $T$ temporally ordered clips following~\cite{55}, yielding synchronised face, body, audio, and text streams for participants $A$ and $B$ (Fig.~\ref{Fig: 2}(a)). Modality-specific encoders then extract low- and high-level features~\cite{32,33,34} (Fig.~\ref{Fig: 2}(b)). Homogeneous and heterogeneous cross-participant interactions are constructed separately at each semantic level (Fig.~\ref{Fig: 2}(c)). MoI-Former subsequently performs relation-aware sparse reasoning over these interactions and aggregates the selected evidence into a relationship representation (Fig.~\ref{Fig: 2}(d)). Finally, HC-Loss decomposes recognition into hierarchy-aware conditional decisions and is jointly optimised with the standard classification objective (Fig.~\ref{Fig: 2}(e)).

\subsection{Feature Extraction}
\subsubsection{Facial Feature Extraction}
For participant $p\in\{A,B\}$ in clip $\mathcal{V}_{t}$, the sampled facial frames are denoted as
\begin{equation}
    \mathfrak{f}_{p,t}^{\mathrm{face}}
    =
    \left(
    \mathfrak{f}_{p,t,1}^{\mathrm{face}},
    \ldots,
    \mathfrak{f}_{p,t,K}^{\mathrm{face}}
    \right),
\end{equation}
where $K$ denotes the number of frames. A low-level facial encoder extracts geometric facial cues, including landmarks and head pose:
\begin{equation}
    \left(
    \mathbf{L}_{p,t},
    \mathbf{H}_{p,t}
    \right)
    =
    \mathcal{E}_{\mathrm{face}}^{\mathrm{low}}
    \left(
    \mathfrak{f}_{p,t}^{\mathrm{face}}
    \right).
\end{equation}

To capture high-level facial behaviours and affective states, we employ separate encoders to extract facial action units, gaze directions, valence-arousal features, and facial appearance:
\begin{equation}
    \begin{aligned}
    \left(
    \mathbf{U}_{p,t},
    \mathbf{G}_{p,t}
    \right)
    &=
    \mathcal{E}_{\mathrm{face}}^{\mathrm{high}}
    \left(
    \mathfrak{f}_{p,t}^{\mathrm{face}}
    \right),\\
    \mathbf{A}_{p,t}^{\mathrm{VA}}
    &=
    \mathcal{E}_{\mathrm{face}}^{\mathrm{aff}}
    \left(
    \mathfrak{f}_{p,t}^{\mathrm{face}}
    \right),\\
    \mathbf{v}_{p,t}^{\mathrm{face}}
    &=
    \mathcal{E}_{\mathrm{face}}^{\mathrm{vis}}
    \left(
    \mathfrak{f}_{p,t}^{\mathrm{face}}
    \right).
    \end{aligned}
\end{equation}

The frame-wise facial descriptors are aggregated into clip-level representations using statistical temporal pooling:
\begin{equation}
    \mathcal{S}(\mathbf{X})
    =
    \boldsymbol{\mu}(\mathbf{X})
    \Vert
    \boldsymbol{\sigma}(\mathbf{X})
    \Vert
    \boldsymbol{\delta}(\mathbf{X}),
    \label{eq:statistical_pooling}
\end{equation}
where $\boldsymbol{\mu}(\cdot)$, $\boldsymbol{\sigma}(\cdot)$, and $\boldsymbol{\delta}(\cdot)$ represent the temporal mean, standard deviation, and first-order variation, respectively. Accordingly,
\begin{equation}
    \widetilde{\mathbf{Z}}_{p,t}
    =
    \mathcal{S}
    \left(
    \mathbf{Z}_{p,t}
    \right),
    \qquad
    \mathbf{Z}
    \in
    \left\{
    \mathbf{L},
    \mathbf{H},
    \mathbf{U},
    \mathbf{G},
    \mathbf{A}^{\mathrm{VA}}
    \right\}.
\end{equation}

Finally, the aggregated facial cues are organised according to their semantic levels:
\begin{equation}
    \begin{aligned}
    \mathbf{F}_{\mathrm{f},p,t}^{\mathrm{low}}
    &=
    \phi_{\mathrm{face}}^{\mathrm{low}}
    \left(
    \widetilde{\mathbf{L}}_{p,t}
    \Vert
    \widetilde{\mathbf{H}}_{p,t}
    \right),
    \\
    \mathbf{F}_{\mathrm{f},p,t}^{\mathrm{high}}
    &=
    \phi_{\mathrm{face}}^{\mathrm{high}}
    \left(
    \widetilde{\mathbf{U}}_{p,t}
    \Vert
    \widetilde{\mathbf{G}}_{p,t}
    \Vert
    \widetilde{\mathbf{A}}_{p,t}^{\mathrm{VA}}
    \Vert
    \mathbf{v}_{p,t}^{\mathrm{face}}
    \right),
    \end{aligned}
\end{equation}
where $\phi_{\mathrm{face}}^{\mathrm{low}}(\cdot)$ and $\phi_{\mathrm{face}}^{\mathrm{high}}(\cdot)$ denote the corresponding projection functions.

\subsubsection{Body Feature Extraction}
For participant $p\in\{A,B\}$ in clip $\mathcal{V}_{t}$, we extract frame-wise body keypoints using a pose encoder:
\begin{equation}
    \mathbf{P}_{p,t}
    =
    \mathcal{E}_{\mathrm{body}}^{\mathrm{low}}
    \left(
    \mathfrak{f}_{p,t}^{\mathrm{body}}
    \right).
\end{equation}
While keypoints describe the geometric body configuration, they do not explicitly capture expressive postural states. Therefore, we further derive body openness and orientation cues from the pose sequence.

Body openness characterises the spatial expansion of the torso and limbs. For the $r$-th frame, the torso length
\begin{equation}
    \lambda_{r}
    =
    \max
    \left(
    \left\|
    \mathbf{n}_{r}-\mathbf{m}_{r}
    \right\|_{2},
    \epsilon
    \right)
\end{equation}
is used as a scale reference, where $\mathbf{n}_{r}$ and $\mathbf{m}_{r}$ denote the shoulder and hip midpoints, respectively. Based on the body configuration, we compute the openness descriptor:
\begin{equation}
    \mathbf{o}_{p,t,r}
    =
    \left[
    s_{r},
    h_{r},
    a_{r}^{L},
    a_{r}^{R},
    w_{r},
    e_{r},
    l_{r},
    A_{r}
    \right]^{\mathsf{T}},
\end{equation}
where $s_r$ and $h_r$ denote the shoulder and hip spans, $a_r^{L}$ and $a_r^{R}$ represent the left and right arm extensions measured from the shoulder midpoint to the corresponding wrists, and $w_r$, $e_r$, and $l_r$ indicate the bilateral distances between wrists, elbows, and ankles, respectively. $A_r$ denotes the projected torso area enclosed by the shoulder and hip joints. All geometric measurements are normalised by $\lambda_r$ to reduce variations caused by body scale and camera distance.

Body orientation is represented by the relative directions of the shoulder, hip, and torso axes. Specifically, let $\mathbf{u}_{r}^{s}$, $\mathbf{u}_{r}^{h}$, and $\mathbf{u}_{r}^{t}$ denote the normalised shoulder, hip, and torso directions, respectively. Their concatenation forms the orientation descriptor:
\begin{equation}
    \mathbf{r}_{p,t,r}
    =
    \mathbf{u}_{r}^{s}
    \Vert
    \mathbf{u}_{r}^{h}
    \Vert
    \mathbf{u}_{r}^{t}.
\end{equation}
The frame-wise descriptors are stacked to obtain the openness sequence $\mathbf{O}_{p,t}$ and orientation sequence $\mathbf{R}_{p,t}$.

Following the facial stream, statistical temporal pooling in Eq.~\ref{eq:statistical_pooling} is applied to aggregate frame-wise body descriptors:
\begin{equation}
    \begin{aligned}
    \widetilde{\mathbf{P}}_{p,t}
    &=
    \mathcal{S}
    \left(
    \mathbf{P}_{p,t}
    \right),\quad
    \widetilde{\mathbf{O}}_{p,t}
    =
    \mathcal{S}
    \left(
    \mathbf{O}_{p,t}
    \right),\\
    \widetilde{\mathbf{R}}_{p,t}
    &=
    \mathcal{S}
    \left(
    \mathbf{R}_{p,t}
    \right).
    \end{aligned}
\end{equation}
Meanwhile, a visual encoder extracts clip-level body appearance:
\begin{equation}
    \mathbf{v}_{p,t}^{\mathrm{body}}
    =
    \mathcal{E}_{\mathrm{body}}^{\mathrm{vis}}
    \left(
    \mathfrak{f}_{p,t}^{\mathrm{body}}
    \right).
\end{equation}

Finally, the pooled keypoints constitute the low-level body representation, while openness, orientation, and appearance cues are projected into the high-level representation:
\begin{equation}
    \begin{aligned}
    \mathbf{F}_{\mathrm{b},p,t}^{\mathrm{low}}
    &=
    \phi_{\mathrm{body}}^{\mathrm{low}}
    \left(
    \widetilde{\mathbf{P}}_{p,t}
    \right),\\
    \mathbf{F}_{\mathrm{b},p,t}^{\mathrm{high}}
    &=
    \phi_{\mathrm{body}}^{\mathrm{high}}
    \left(
    \widetilde{\mathbf{O}}_{p,t}
    \Vert
    \widetilde{\mathbf{R}}_{p,t}
    \Vert
    \mathbf{v}_{p,t}^{\mathrm{body}}
    \right).
    \end{aligned}
\end{equation}

\subsubsection{Audio Feature Extraction}
For participant $p\in\{A,B\}$ in clip $\mathcal{V}_{t}$, a low-level acoustic encoder extracts spectral and prosodic descriptors, including MFCC, fundamental frequency (F0), and energy:
\begin{equation}
    \left(
    \mathbf{C}_{p,t},
    \mathbf{Q}_{p,t},
    \mathbf{E}_{p,t}
    \right)
    =
    \mathcal{E}_{\mathrm{audio}}^{\mathrm{low}}
    \left(
    \mathfrak{f}_{p,t}^{\mathrm{audio}}
    \right),
\end{equation}
where these descriptors capture complementary acoustic characteristics of the speech signal.

To capture higher-level vocal and temporal cues, a high-level audio encoder extracts acoustic emotion, speech dynamics, and spectral voice characteristics:
\begin{equation}
    \left(
    \mathbf{A}_{p,t}^{\mathrm{emo}},
    \mathbf{T}_{p,t}^{\mathrm{dyn}},
    \mathbf{S}_{p,t}^{\mathrm{spec}}
    \right)
    =
    \mathcal{E}_{\mathrm{audio}}^{\mathrm{high}}
    \left(
    \mathfrak{f}_{p,t}^{\mathrm{audio}}
    \right).
\end{equation}
Specifically, the temporal dynamics descriptor models speech activity and pause patterns:
\begin{equation}
    \mathbf{T}_{p,t}^{\mathrm{dyn}}
    =
    \left(
    d_{p,t}^{\mathrm{speech}},
    d_{p,t}^{\mathrm{silence}},
    d_{p,t}^{\mathrm{pause}},
    \nu_{p,t}^{\mathrm{pause}}
    \right),
\end{equation}
while the spectral descriptor captures voice quality variations:
\begin{equation}
    \mathbf{S}_{p,t}^{\mathrm{spec}}
    =
    \left(
    r_{p,t}^{\alpha},
    h_{p,t}^{\mathrm{Hamm}},
    s_{p,t}^{\mathrm{flux}}
    \right).
\end{equation}

The low- and high-level audio representations are then obtained by projecting the corresponding acoustic cues:
\begin{equation}
    \begin{aligned}
    \mathbf{F}_{\mathrm{a},p,t}^{\mathrm{low}}
    &=
    \phi_{\mathrm{audio}}^{\mathrm{low}}
    \left(
    \mathbf{C}_{p,t}
    \Vert
    \mathbf{Q}_{p,t}
    \Vert
    \mathbf{E}_{p,t}
    \right),
    \\
    \mathbf{F}_{\mathrm{a},p,t}^{\mathrm{high}}
    &=
    \phi_{\mathrm{audio}}^{\mathrm{high}}
    \left(
    \mathbf{A}_{p,t}^{\mathrm{emo}}
    \Vert
    \mathbf{T}_{p,t}^{\mathrm{dyn}}
    \Vert
    \mathbf{S}_{p,t}^{\mathrm{spec}}
    \right),
    \end{aligned}
\end{equation}
where $\phi_{\mathrm{audio}}^{\mathrm{low}}(\cdot)$ and $\phi_{\mathrm{audio}}^{\mathrm{high}}(\cdot)$ denote the corresponding projection functions.

\subsubsection{Text Feature Extraction}
For the textual stream $\mathfrak{f}_{p,t}^{\mathrm{text}}$, we use a low-level text encoder $\mathcal{E}_{\mathrm{text}}^{\mathrm{low}}$ to capture token-level linguistic representations $\mathbf{F}_{\mathrm{t},p,t}^{\mathrm{low}}$, while a high-level encoder $\mathcal{E}_{\mathrm{text}}^{\mathrm{high}}$ extracts textual sentiment $\mathbf{F}_{\mathrm{t},p,t}^{\mathrm{high}}$:
\begin{equation}
    \begin{aligned}
    \mathbf{F}_{\mathrm{t},p,t}^{\mathrm{low}}
    &=
    \mathcal{E}_{\mathrm{text}}^{\mathrm{low}}
    \left(
    \mathfrak{f}_{p,t}^{\mathrm{text}}
    \right),
    \\
    \mathbf{F}_{\mathrm{t},p,t}^{\mathrm{high}}
    &=
    \mathcal{E}_{\mathrm{text}}^{\mathrm{high}}
    \left(
    \mathfrak{f}_{p,t}^{\mathrm{text}}
    \right).
    \end{aligned}
\end{equation}

\subsection{Interaction Feature Construction}
\begin{figure}[ht]
    \vspace{-0.5em}
    \setlength{\abovecaptionskip}{0.cm}
    \setlength{\belowcaptionskip}{-0.cm}
    \centering
    \includegraphics[width=3.0in]{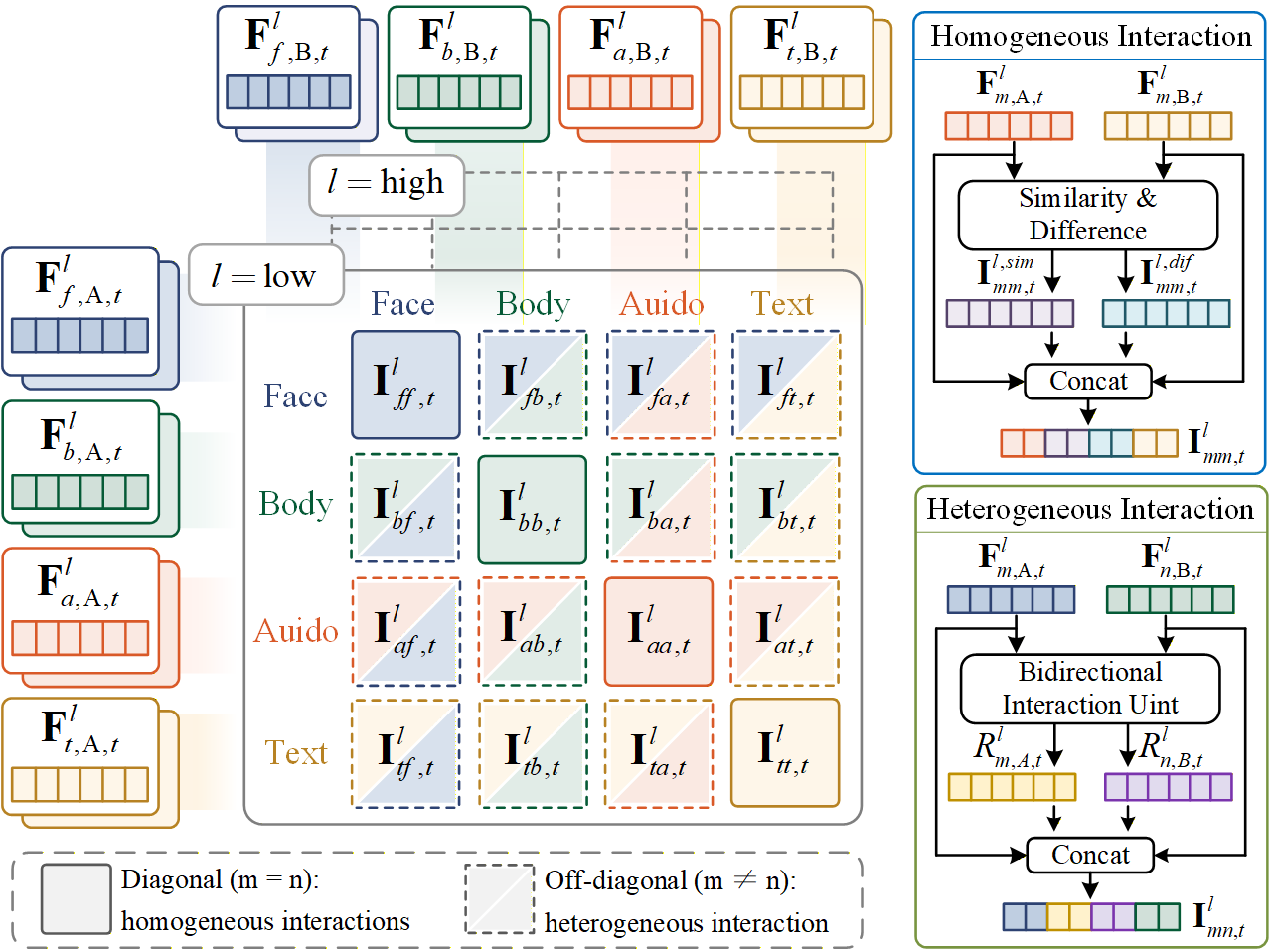}
    \caption{Interaction Feature Construction. At each feature level, the face, body, audio, and text representations of participants $A$ and $B$ are paired to form a $4\times4$ interaction matrix.
    \label{Fig: 3}}
    \vspace{-0.5em}
\end{figure}

We construct interpersonal interactions separately at each feature level $l\in\{\mathrm{low},\mathrm{high}\}$, as illustrated in Fig.~\ref{Fig: 3}. Given participant features from different modalities, we model both homogeneous interactions within the same modality space and heterogeneous interactions across modality spaces.

For homogeneous interactions, the behavioural correspondence between participants is captured through similarity and discrepancy cues. Specifically, for modality $m$, we define
\begin{equation}
    \mathbf{I}_{mm,t}^{l,\mathrm{sim}}
    =
    \frac{
        \mathbf{F}_{m,A,t}^{l}
        \odot
        \mathbf{F}_{m,B,t}^{l}
    }{
        \left\|
        \mathbf{F}_{m,A,t}^{l}
        \right\|_{2}
        \left\|
        \mathbf{F}_{m,B,t}^{l}
        \right\|_{2}
    },
\end{equation}
and
\begin{equation}
    \mathbf{I}_{mm,t}^{l,\mathrm{dif}}
    =
    \left|
        \mathbf{F}_{m,A,t}^{l}
        -
        \mathbf{F}_{m,B,t}^{l}
    \right|,
\end{equation}
where the similarity term describes behavioural consistency, while the discrepancy term captures individual differences. The homogeneous interaction token is then formulated as
\begin{equation}
    \mathbf{I}_{mm,t}^{l}
    =
    \mathbf{F}_{m,A,t}^{l}
    \Vert
    \mathbf{F}_{m,B,t}^{l}
    \Vert
    \mathbf{I}_{mm,t}^{l,\mathrm{sim}}
    \Vert
    \mathbf{I}_{mm,t}^{l,\mathrm{dif}}.
\end{equation}

For heterogeneous interactions, features from different modalities cannot be directly compared. We therefore introduce a bidirectional interaction unit to model cross-modal dependencies while preserving the modality-specific perspectives. For an ordered modality pair $(m,n)$ with $m\neq n$, the interaction components are generated as
\begin{equation}
    \mathbf{R}_{m,A,t}^{l}
    =
    \psi_{m}^{l}
    \left(
        \mathbf{F}_{m,A,t}^{l}
        \Vert
        \mathbf{F}_{n,B,t}^{l}
    \right),
\end{equation}
\begin{equation}
    \mathbf{R}_{n,B,t}^{l}
    =
    \psi_{n}^{l}
    \left(
        \mathbf{F}_{n,B,t}^{l}
        \Vert
        \mathbf{F}_{m,A,t}^{l}
    \right),
\end{equation}
where $\psi_m^l(\cdot)$ and $\psi_n^l(\cdot)$ denote lightweight mapping functions. The heterogeneous interaction token is constructed as
\begin{equation}
    \mathbf{I}_{mn,t}^{l}
    =
    \mathbf{F}_{m,A,t}^{l}
    \Vert
    \mathbf{F}_{n,B,t}^{l}
    \Vert
    \mathbf{R}_{m,A,t}^{l}
    \Vert
    \mathbf{R}_{n,B,t}^{l},
    \qquad m\neq n .
\end{equation}

\begin{figure*}[t!] 
\vspace{-0.5em}
\setlength{\abovecaptionskip}{0.cm}     
\setlength{\belowcaptionskip}{-0.cm} 
\centering     
\includegraphics[width=6.0in]{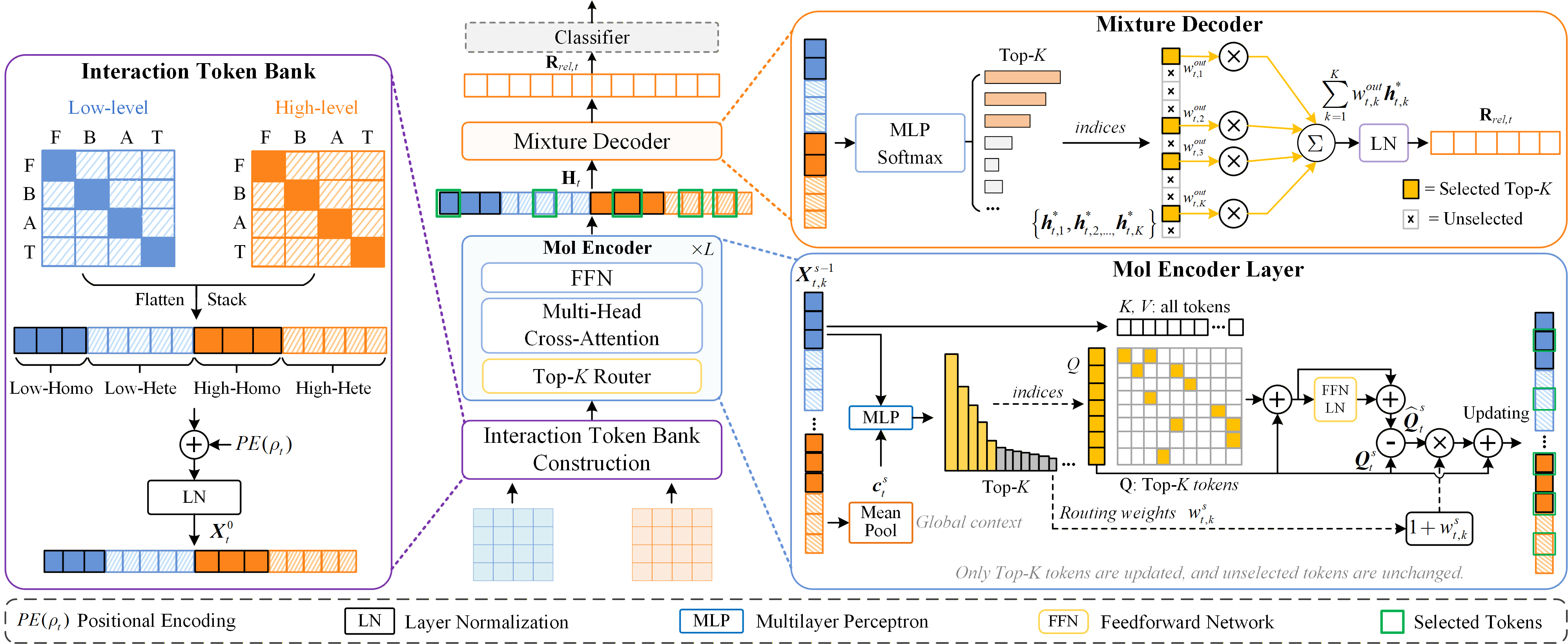} 
    \caption{Overview of the proposed MoI-Former. Low- and high-level homogeneous and heterogeneous interactions are organised into a unified interaction token bank. Relation-aware sparse routing selects the Top-$K$ informative tokens for contextual refinement, while the complete interaction bank remains available as keys and values. Updated tokens are written back to their original positions, and routing is progressively refined across layers. The decoder aggregates the selected interaction tokens into the final relationship representation. \label{Fig: 4}} 
    \vspace{-1.0em} 
\end{figure*}

\subsection{Mixture-of-Interaction Transformer}
Different relationships rely on distinct combinations of facial, bodily, acoustic, and linguistic interactions. Directly aggregating all interaction cues may introduce irrelevant evidence. To address this issue, we propose MoI-Former, which regards interaction patterns as specialised experts and performs relation-aware sparse routing to identify relationship-relevant evidence, as shown in Fig.~\ref{Fig: 4}.

For clip $\mathcal{V}_{t}$, low- and high-level interaction matrices are flattened and concatenated into an interaction token bank:
\begin{equation}
    \mathbf{I}_{t}
    =
    \left[
    \mathbf{I}_{t}^{\mathrm{low}};
    \mathbf{I}_{t}^{\mathrm{high}}
    \right]
    =
    \left[
    \mathbf{i}_{t,1},
    \ldots,
    \mathbf{i}_{t,N}
    \right]^{\mathsf{T}}
    \in
    \mathbb{R}^{N\times d_I},
    \label{eq:moi_token_bank}
\end{equation}
where $N=2M^2=32$ for $M=4$ modalities and $d_I=4d$. The $32$ tokens represent ordered pairwise interactions across two semantic levels and four modalities. They provide a compact and interpretable interaction basis rather than an exhaustive enumeration of all higher-order or temporal dependencies. Each token represents an ordered cross-participant interaction in a specific semantic level. To preserve temporal information, the normalised temporal position
$\rho_t=(t-1)/\max(T-1,1)$
is encoded and added to all interaction tokens:
\begin{equation}
    \mathbf{X}_{t}^{0}
    =
    \operatorname{LN}
    \left(
    \mathbf{I}_{t}
    +
    \mathbf{1}_{N}
    \operatorname{PE}(\rho_t)^{\mathsf{T}}
    \right).
    \label{eq:moi_input}
\end{equation}

At each encoder layer, MoI-Former dynamically selects informative interactions according to the current relational context. Given $\mathbf{X}_{t}^{s-1}$, a global interaction context is first obtained by mean pooling:
\begin{equation}
    \mathbf{c}_{t}^{s}
    =
    \operatorname{MeanPool}
    \left(
    \mathbf{X}_{t}^{s-1}
    \right).
\end{equation}
Each token is then scored by a context-aware router:
\begin{equation}
    a_{t,k}^{s}
    =
    \operatorname{MLP}_{\theta}^{s}
    \left(
    \mathbf{x}_{t,k}^{s-1}
    \Vert
    \mathbf{c}_{t}^{s}
    \right),
    \qquad
    k=1,\ldots,N ,
    \label{eq:moi_relevance_score}
\end{equation}
and normalised as
\begin{equation}
    p_{t,k}^{s}
    =
    \operatorname{Softmax}_{k}
    \left(
    \frac{
    a_{t,k}^{s}+\epsilon_{t,k}^{s}
    }{\tau}
    \right),
    \label{eq:moi_routing_probability}
\end{equation}
where Gaussian noise $\epsilon_{t,k}^{s}$ is used only during training to encourage exploration.

The Top-$K$ interactions are selected according to the routing probabilities:
\begin{equation}
    \{
    \mathbf{q}_{t,1}^{s},
    \ldots,
    \mathbf{q}_{t,K}^{s}
    \}
    =
    \operatorname{TopK}
    \left(
    \{
    (\mathbf{x}_{t,k}^{s-1},p_{t,k}^{s})
    \}_{k=1}^{N}
    \right),
    \label{eq:moi_topk_selection}
\end{equation}
and their weights are renormalised:
\begin{equation}
    w_{t,k}^{s}
    =
    \frac{
    p(\mathbf{q}_{t,k}^{s})
    }{
    \sum_{j=1}^{K}p(\mathbf{q}_{t,j}^{s})
    } .
    \label{eq:moi_routing_weight}
\end{equation}

Unlike conventional sparse selection that discards inactive interactions, MoI-Former retains the complete interaction bank as contextual memory. The selected tokens serve as queries, while all interactions provide keys and values:
\begin{equation}
    \begin{aligned}
    \mathbf{Y}_{t}^{s}
    &=
    \mathbf{Q}_{t}^{s}
    +
    \operatorname{MCA}
    (
    \operatorname{LN}(\mathbf{Q}_{t}^{s}),
    \operatorname{LN}(\mathbf{X}_{t}^{s-1}),
    \operatorname{LN}(\mathbf{X}_{t}^{s-1})
    ),\\
    \widehat{\mathbf{Q}}_{t}^{s}
    &=
    \mathbf{Y}_{t}^{s}
    +
    \operatorname{FFN}
    (
    \operatorname{LN}(\mathbf{Y}_{t}^{s})
    ).
    \end{aligned}
    \label{eq:moi_sparse_encoding}
\end{equation}
where $\operatorname{MCA}$ is the multi-head cross-attention operator. The selected interactions are further updated through a routing-aware gating mechanism, where the contribution of each token is updated according to its routing weight:
\begin{equation}
    \mathbf{q}_{t,k}^{s,\mathrm{out}}
    =
    \mathbf{q}_{t,k}^{s}
    +
    (1+w_{t,k}^{s})
    \left(
    \widehat{\mathbf{q}}_{t,k}^{s}
    -
    \mathbf{q}_{t,k}^{s}
    \right),
    \label{eq:moi_update}
\end{equation}
where interactions with higher routing weights receive stronger contextual updates. The updated tokens are then written back to their original positions, whereas unselected tokens remain unchanged. Since routing is recomputed at each encoder layer, the activated interaction subset can evolve progressively during relationship reasoning.

After $L$ encoder layers, the contextualised interaction bank $\mathbf{H}_{t}$ is decoded by selecting the final Top-$K$ interactions and aggregating them with routing weights:
\begin{equation}
\mathbf{R}_{\mathrm{rel},t}
=
\operatorname{LN}
\left(
\sum_{k=1}^{K}
w_{t,k}^{\mathrm{out}}
\mathbf{h}_{t,k}^{*}
\right).
\label{eq:moi_relation}
\end{equation}
The resulting representation $\mathbf{R}_{\mathrm{rel},t}$ is used for subsequent relationship recognition.

\subsection{Relationship Classification with HC-Loss}
As shown in Fig.~\ref{Fig: 2}(e), dyadic relationships exhibit an inherent semantic hierarchy rather than independent class structures. Conventional flat classification ignores such hierarchical dependencies and treats all categories equally. To incorporate this structure, we propose Hierarchical Conditional Loss (HC-Loss), which decomposes relationship recognition into a sequence of hierarchy-aware binary decisions

\textit{Hierarchy Definition.}
Let $\mathcal{Y}=\{1,\ldots,C\}$ denote the relationship classes. We represent their semantic organisation as a binary tree with $D$ internal nodes. Each node $d$ corresponds to a subset $\mathcal{C}_{d}$ and splits it into two child subsets:
\begin{equation}
    \mathcal{C}_{d}
    =
    \mathcal{C}_{d}^{0}
    \cup
    \mathcal{C}_{d}^{1},
    \qquad
    \mathcal{C}_{d}^{0}
    \cap
    \mathcal{C}_{d}^{1}
    =
    \varnothing .
    \label{eq:relation_hierarchy}
\end{equation}
For a sample with label $y_i$, only the decision nodes along its root-to-leaf path $\mathcal{P}(y_i)$ are involved. The branch indicator is denoted as $b_d(y_i)\in\{0,1\}$, and the samples participating in node $d$ are
$\mathcal{S}_d=\{i|y_i\in\mathcal{C}_d\}$.

\textit{Hierarchical Conditional Probability.}
Given the relationship representation $\mathbf{R}_{\mathrm{rel},i}$, the classifier outputs logits
$\mathbf{z}_{i}=f_{\mathrm{cls}}(\mathbf{R}_{\mathrm{rel},i})$.
Instead of independently optimising class logits, HC-Loss aggregates class evidence within each branch. The score of branch $b$ at node $d$ is computed as
\begin{equation}
    s_{i,d}^{b}
    =
    \log
    \left(
    \sum_{c\in\mathcal{C}_{d}^{b}}
    \exp(z_{i,c})
    \right).
    \label{eq:branch_score}
\end{equation}
The conditional probability of selecting branch $b$ is obtained as
\begin{equation}
    q_{i,d}^{\,b}
    =
    \begin{cases}
        \sigma(\eta_{i,d}), & b=1,\\
        1-\sigma(\eta_{i,d}), & b=0,
    \end{cases}
    \qquad
    \eta_{i,d}=s_{i,d}^{1}-s_{i,d}^{0}.
    \label{eq:conditional_prob}
\end{equation}

\textit{Training Objective.}
Each hierarchical decision is supervised by
\begin{equation}
    \mathcal{L}_{d}
    =
    -\frac{1}{|\mathcal{S}_d|}
    \sum_{i\in\mathcal{S}_d}
    \log q_{i,d}^{b_d(y_i)} .
    \label{eq:node_loss}
\end{equation}
The hierarchical loss aggregates all decision nodes:
\begin{equation}
    \mathcal{L}_{\mathrm{hier}}
    =
    \frac{
    \sum_{d=1}^{D}w_d\mathcal{L}_d
    }{
    \sum_{d=1}^{D}w_d
    },
\end{equation}
where $w_d$ controls the contribution of each decision node.

To preserve direct class-level supervision, we also employ the standard classification loss:
\begin{equation}
    \mathcal{L}_{\mathrm{cls}}
    =
    -\frac{1}{N}
    \sum_{i=1}^{N}
    \log p_{i,y_i},
    \qquad
    \mathbf{p}_{i}
    =
    \operatorname{Softmax}(\mathbf{z}_{i}).
    \label{eq:cls_loss}
\end{equation}
The final objective combines cross-entropy loss and hierarchical conditional loss:
\begin{equation}
    \mathcal{L}
    =
    \mathcal{L}_{\mathrm{cls}}
    +
    \lambda
    \mathcal{L}_{\mathrm{hier}},
    \label{eq:total_loss}
\end{equation}
where $\lambda$ balances the two objectives. This formulation preserves fine-grained class discrimination while encouraging predictions to follow the semantic organisation of relationship categories.

\section{Evaluation Setup}  
\subsection{Datasets}  
We evaluate our method on NoXi~\cite{49}, UDIVA~\cite{48}, and Seamless Interaction (SeamInt)~\cite{56}. NoXi comprises 168 videos spanning four relationship classes: \textit{Stranger} (Str.), \textit{Acquaintance} (Acq.), \textit{Friend} (Fri.), and \textit{Very Good Friend} (Vgf.). The dataset provides annotations from two participant perspectives, referred to as NoXi-I and NoXi-J. UDIVA contains 1,160 videos and defines a binary relationship-recognition task with the classes \textit{Known} (Kno.) and \textit{Unknown} (Unk.). SeamInt is substantially larger, including more than 4,000 hours of face-to-face interactions involving over 4,000 participants, with a total data volume of approximately 27~TB. Owing to computational and storage constraints, we randomly sampled 2,738 videos from SeamInt for our experiments, covering five relationship classes: Str., Acq., Fri., Fam., and Cou. Following the clip-construction protocol of~\cite{55}, each video is partitioned into temporally ordered clips, which constitute the clip-level datasets used for training and evaluation. 



\subsection{Implementation Details}
\subsubsection{Feature Extraction}
As summarised in Table~\ref{Table: 1}, modality-specific encoders extract complementary low- and high-level features from facial, bodily, acoustic, and textual signals. Low-level features describe fundamental geometric, acoustic, and semantic cues, whereas high-level features capture behavioural, affective, and appearance-related information. These features are temporally aggregated at the clip level and projected into a unified representation space for dyadic interaction modelling. The pre-trained encoders remain frozen, while the projection and interaction reasoning modules are optimised end-to-end.

\begin{table*}[t!]\footnotesize
\vspace{-0.5em}
\caption{Pre-trained models used for feature extraction.}
\vspace{-1.0em}
\centering
\def\temptablewidth{0.2\textwidth}
\tabcolsep=0.25cm
\renewcommand\arraystretch{1.1}
\begin{threeparttable}
    \begin{tabular}{c|cc|cc}
    \hline
    \multirow{2}{*}{Modality} & \multicolumn{2}{c|}{Low-level Features} & \multicolumn{2}{c}{High-level Features} \\ \cline{2-5}
                              & \multicolumn{1}{c|}{Features} & Extractor & \multicolumn{1}{c|}{Features} & Extractor \\ \hline
    Face &
      \multicolumn{1}{c|}{
      \begin{tabular}[c]{@{}c@{}}
      Facial landmarks,\\
      Head pose coordinates
      \end{tabular}} &
      OpenFace 2.0~\cite{67} &
      \multicolumn{1}{c|}{
      \begin{tabular}[c]{@{}c@{}}
      Facial action units,\\
      Eye gaze directions,\\
      Valence and Arousal,\\
      Visual appearance
      \end{tabular}} &
      \begin{tabular}[c]{@{}c@{}}
      OpenFace 2.0~\cite{67},\\
      EmoNet~\cite{77},\\
      TimeSformer~\cite{68}
      \end{tabular} \\ \hline
    Body &
      \multicolumn{1}{c|}{Body pose coordinates} &
      Keypoint R-CNN~\cite{69} &
      \multicolumn{1}{c|}{
      \begin{tabular}[c]{@{}c@{}}
      Body openness and orientation,\\
      Visual appearance
      \end{tabular}} &
      \begin{tabular}[c]{@{}c@{}}
      Pose-based geometric modelling,\\
      TimeSformer~\cite{68}
      \end{tabular} \\ \hline
    Audio &
      \multicolumn{1}{c|}{
      \begin{tabular}[c]{@{}c@{}}
      20 MFCC, F0, RMS
      \end{tabular}} &
      Librosa~\cite{70} &
      \multicolumn{1}{c|}{
      \begin{tabular}[c]{@{}c@{}}
      Acoustic emotion, Spectral flux\\
      Speech/Silence duration,\\
      Pause duration/frequency,\\
      Alpha ratio, Hammarberg index
      \end{tabular}} &
      \begin{tabular}[c]{@{}c@{}}
      HuBERT-Large~\cite{71},\\
      Librosa~\cite{70},\\
      openSMILE~\cite{78}
      \end{tabular} \\ \hline
    Text &
      \multicolumn{1}{c|}{Semantic embedding} &
      BERT-base~\cite{72} &
      \multicolumn{1}{c|}{Sentiment of the speech} &
      DeBERTa-base~\cite{73} \\ \hline
    \end{tabular}
    \footnotesize
\end{threeparttable}
\label{Table: 1}
\vspace{-1.0em}
\end{table*}

\subsubsection{Experimental Setup}
All experiments are implemented in PyTorch and conducted on a server equipped with an Intel(R) Xeon(R) Gold 6330 CPU @ 2.00\,GHz, 256\,GB RAM, and an NVIDIA A100 80\,GB PCIe GPU. Following~\cite{55}, we adopt 3-fold cross-validation with non-overlapping dyadic video partitions, ensuring that clips from the same interaction are not shared between training and testing sets. We train the model for 100 epochs using AdamW with a batch size of 64. The initial learning rate is set to $1\times10^{-4}$ with a weight decay of $5\times10^{-4}$, and is decayed by a factor of 10 every 15 epochs.

For MoI-Former, interaction tokens are projected into a 2048-dimensional latent space and processed by two Transformer encoder layers with 8 attention heads, followed by the interaction decoder. The sparse router selects the top-$K=5$ interaction tokens at each routing stage, and the dropout rate is set to 0.3.

\subsubsection{Evaluation Metrics}
Considering the class imbalance in social relationship datasets, we follow the evaluation protocol commonly adopted in~\cite{52,55} and report class-wise Recall, Unweighted Average Recall (UAR), and mean Average Precision (mAP) for fair comparison. Class-wise Recall measures recognition performance for individual relationship categories, while UAR gives equal weight to all classes and is therefore robust to imbalanced distributions. Additionally, mAP evaluates the quality of predicted confidence scores across relationship categories.

\subsection{Baseline Models}
We compare HiRe-DRR with representative image-based, video-based, and multimodal relationship recognition methods. For image-based baselines, we adapt them to the video setting by temporally aggregating frame-level visual features within each clip to obtain person-level representations, which are then processed by their original relation reasoning modules. Audio and textual modalities are excluded to maintain their original modality assumptions.

\begin{itemize}
    \item \textbf{GR2N}~\cite{22} performs image-based social relationship reasoning by constructing pairwise graph structures. (\textit{\textbf{Image-bsed method}})

    \item \textbf{HGM}~\cite{17} extends pairwise graph reasoning to higher-order interaction modelling by employing hypergraphs. (\textit{\textbf{Image-bsed method}})

    \item \textbf{PGCN}~\cite{18} models video-based social relationships through pyramid graph convolution. (\textit{\textbf{Video-bsed method}})

    \item \textbf{LIReC}~\cite{19} is a multimodal framework that jointly models visual and textual cues to infer character social relationships. (\textit{\textbf{Multimodal-bsed method}})

    \item \textbf{Cumulative Transformer (CT)}~\cite{29} is Transformer architecture that captures long-term relationship dynamics by progressively accumulating historical interaction representations. (\textit{\textbf{Multimodal-bsed method}})

    \item \textbf{AsyReC}~\cite{55} is multimodal framework that integrates facial, bodily, acoustic, and textual cues, while modelling directional interpersonal dependencies and temporal interaction patterns. (\textit{\textbf{Multimodal-bsed method}})
\end{itemize}

\section{Experiments and Results}
\subsection{Comparative Analyses}
Table~\ref{Table: 2} compare HiRe-DRR with representative image-, video-, and multimodal relationship recognition methods. HiRe-DRR achieves the best overall UAR/mUAR and mAP on all three datasets, indicating the benefit of jointly modelling feature levels, cross-participant interactions, and relationship hierarchy.

\begin{table*}[t!]
\footnotesize
\caption{Comparison with representative baselines on three dyadic relationship benchmarks.}
\label{Table: 2}
\vspace{-0.8em}

\centering


\textbf{(a) NoXi}
\vspace{0.3em}

\setlength{\tabcolsep}{0.25cm}
\renewcommand{\arraystretch}{1.2}

\begin{threeparttable}
\begin{tabular}{l|cccc|c|cccc|c|cc}
\hline
\multirow{2}{*}{Methods}
& \multicolumn{5}{c|}{NoXi-I}
& \multicolumn{5}{c|}{NoXi-J}
& \multirow{2}{*}{mUAR}
& \multirow{2}{*}{mAP} \\ \cline{2-11}
& Str. & Acq. & Fri. & Vgf. & UAR
& Str. & Acq. & Fri. & Vgf. & UAR
& & \\ \hline

GR2N~\cite{22}
& 65.3 & 49.0 & 17.7 & 40.7 & 43.2\std{8.6}
& 57.8 & 59.9 & 9.1 & 28.1 & 39.7\std{10.5}
& 40.9\std{8.9} & 48.3\std{9.9} \\

HGM~\cite{17}
& 65.1 & 38.9 & 34.8 & 48.3 & 46.7\std{4.5}
& 66.2 & 49.9 & 15.2 & 22.6 & 38.5\std{8.6}
& 42.6\std{7.6} & 49.5\std{9.6} \\

PGCN~\cite{18}
& 21.4 & 19.1 & 0.0 & 64.1 & 26.2\std{4.0}
& 19.4 & 21.8 & 50.7 & 17.9 & 27.5\std{7.6}
& 26.8\std{6.6} & 48.9\std{6.9} \\

LIReC~\cite{19}
& 51.6 & 45.4 & 19.3 & 33.5 & 37.0\std{3.1}
& 70.6 & 45.4 & 19.0 & 10.4 & 36.4\std{13.9}
& 36.9\std{9.9} & 49.6\std{8.6} \\

CT~\cite{29}
& 61.6 & 57.2 & 33.2 & 8.6 & 40.2\std{16.3}
& 47.7 & 57.7 & 30.0 & 25.1 & 40.1\std{11.1}
& 40.1\std{13.9} & 51.7\std{7.8} \\

AsyReC~\cite{55}
& 49.6 & 61.9 & 38.4 & 52.8 & 50.7\std{3.6}
& 47.2 & 59.7 & 40.5 & 35.4 & 45.7\std{4.3}
& 48.2\std{6.5} & 52.1\std{10.3} \\

\textbf{HiRe-DRR (ours)}
& 69.3 & 56.9 & 56.3 & 71.8 & \textbf{63.6\std{2.5}}
& 65.4 & 53.7 & 44.1 & 50.8 & \textbf{53.5\std{6.3}}
& \textbf{58.5\std{7.0}} & \textbf{63.1\std{6.1}} \\

\hline
\end{tabular}

\begin{tablenotes}[flushleft]
\scriptsize
\item[] UAR, mUAR, and mAP are reported in \%; the remaining columns report class-wise recall.
Str.: Stranger; Acq.: Acquaintance; Fri.: Friend; Vgf.: Very Good Friend.
\end{tablenotes}
\end{threeparttable}

\vspace{1.0em}


\textbf{(b) UDIVA and SeamInt}
\vspace{0.3em}

\setlength{\tabcolsep}{0.25cm}

\begin{threeparttable}
\begin{tabular}{l|cc|cc|ccccc|cc}
\hline
\multirow{2}{*}{Methods}
& \multicolumn{4}{c|}{UDIVA}
& \multicolumn{7}{c}{SeamInt} \\ \cline{2-12}
& Kno. & Unk. & UAR & mAP
& Str. & Acq. & Fri. & Fam. & Cou. & UAR & mAP \\ \hline

GR2N~\cite{22}
& 50.0 & 53.0 & 51.5\std{5.1} & 53.2\std{6.6}
& 45.0 & 43.4 & 40.9 & 33.2 & 36.7
& 39.8\std{2.1} & 36.1\std{6.2} \\

HGM~\cite{17}
& 59.2 & 55.3 & 57.4\std{5.0} & 59.2\std{7.2}
& 48.4 & 48.1 & 43.1 & 32.7 & 31.5
& 40.7\std{2.5} & 43.0\std{3.9} \\

PGCN~\cite{18}
& 72.3 & 29.9 & 51.1\std{5.2} & 53.9\std{7.2}
& 22.3 & 16.5 & 19.2 & 22.3 & 23.4
& 20.7\std{0.5} & 20.8\std{0.2} \\

LIReC~\cite{19}
& 37.7 & 69.0 & 53.5\std{6.7} & 57.1\std{4.8}
& 42.0 & 32.3 & 35.9 & 40.6 & 62.6
& 42.7\std{1.1} & 42.7\std{0.8} \\

CT~\cite{29}
& 70.3 & 35.7 & 53.0\std{3.7} & 56.2\std{8.4}
& 51.8 & 36.5 & 40.7 & 46.7 & 65.6
& 48.3\std{2.3} & 48.8\std{0.4} \\

AsyReC~\cite{55}
& 65.5 & 53.3 & 59.3\std{6.8} & 63.6\std{7.1}
& 50.7 & 38.5 & 46.6 & 48.9 & 66.8
& 50.3\std{7.9} & 52.2\std{3.6} \\

\textbf{HiRe-DRR (ours)}
& 60.4 & 64.8 & \textbf{62.5\std{5.3}} & \textbf{66.8\std{6.4}}
& 51.6 & 59.4 & 60.2 & 68.6 & 75.1
& \textbf{63.0\std{1.8}} & \textbf{68.5\std{2.6}} \\

\hline
\end{tabular}

\begin{tablenotes}[flushleft]
\scriptsize
\item[] UAR and mAP are reported in \%; the remaining columns report class-wise recall.
Kno.: Known; Unk.: Unknown; Str.: Stranger; Acq.: Acquaintance; Fri.: Friend;
Fam.: Family; Cou.: Couple.
\end{tablenotes}
\end{threeparttable}

\vspace{-1.5em}
\end{table*}

On NoXi, HiRe-DRR achieves UARs of 63.6\% and 53.5\% on NoXi-I and NoXi-J, respectively, yielding an mUAR of 58.5\% and an mAP of 63.1\%. It surpasses the strongest baseline, AsyReC~\cite{55}, by 10.3 and 11.0 percentage points in mUAR and mAP, respectively. Notably, HiRe-DRR obtains the highest \textit{Very Good Friend} recall on both partitions and the highest \textit{Friend} recall on NoXi-I. These results indicate that preserving feature-level semantics and explicitly modelling homogeneous and heterogeneous interactions provide discriminative evidence for distinguishing semantically close relationships. MoI-Former further selects relationship-relevant interaction patterns instead of uniformly fusing all cues, while HC-Loss promotes hierarchy-aware separation between adjacent categories.

On UDIVA, HiRe-DRR achieves 62.5\% UAR and 66.8\% mAP, exceeding AsyReC by 3.2 percentage points on both metrics. It also produces balanced recalls of 60.4\% and 64.8\% for the \textit{Known} and \textit{Unknown} classes, respectively. The smaller gains relative to NoXi and SeamInt may be attributed to UDIVA's seated, upper-body setting, which restricts variations in pose, openness, and orientation and consequently limits the contribution of bodily interaction cues. Nevertheless, the improvements in both overall metrics demonstrate the robustness of HiRe-DRR under restricted behavioural observations.

On SeamInt, HiRe-DRR obtains 63.0\% UAR and 68.5\% mAP, outperforming AsyReC by 12.7 and 16.3 percentage points, respectively. Compared with NoXi and UDIVA, SeamInt contains a broader relationship taxonomy and richer behavioural variations, providing greater scope for exploiting complementary multimodal interactions. In this setting, the adaptive routing of distinct interaction combinations is particularly beneficial, leading to substantial gains over uniform multimodal fusion.

\begin{table}[ht]
\vspace{-0.5em}
\footnotesize
\caption{Ablation results using all modalities.}
\vspace{-0.8em}
\label{Table: 3}
\centering
\setlength{\tabcolsep}{0.35cm}
\renewcommand{\arraystretch}{1.05}
\begin{threeparttable}
\begin{tabular}{lccc}
\toprule
\multirow{2}{*}{Configuration}
& \multicolumn{3}{c}{All modalities} \\ \cmidrule{2-4}
& NoXi & UDIVA & SeamInt \\ 
\midrule
(a) L
& 43.6\std{2.5} & 51.6\std{2.0} & 39.9\std{1.9} \\

(b) H
& 49.0\std{5.6} & 55.9\std{2.0} & 55.3\std{4.2}* \\

(c) LH+MoI
& 55.1\std{5.0} & 65.9\std{4.5}* & 62.4\std{2.4} \\

(d) LH+MoI+HC
& \textbf{63.1}\std{5.6} & \textbf{66.8}\std{6.4} & \textbf{68.5}\std{2.6} \\
\bottomrule
\end{tabular}

\begin{tablenotes}[flushleft]
\scriptsize
\item[] $\ast$ indicates a statistically significant improvement over the preceding configuration ($p<0.05$).
\end{tablenotes}
\end{threeparttable}
\vspace{-1.0em}
\end{table}

\subsection{Ablation Analysis}
\subsubsection{Ablation of the Entire Framework}
To examine the contribution of different components in HiRe-DRR, we consider four configurations:
\begin{itemize}
    \item (a) low-level features only (\textbf{L});
    \item (b) high-level features only (\textbf{H});
    \item (c) joint low- and high-level interaction reasoning (\textbf{LH+MoI});
    \item (d) the complete model with hierarchical supervision (\textbf{LH+MoI+HC}).
\end{itemize}

\begin{table*}[t]
\vspace{-0.5em}
\footnotesize
\caption{Ablation results by modality.}
\vspace{-0.8em}
\label{Table: 4}
\centering
\setlength{\tabcolsep}{0.12cm}
\renewcommand{\arraystretch}{1.20}
\begin{threeparttable}
\begin{tabular}{lcccc|cccc|cccc}
\toprule
\multirow{2}{*}{Configuration}
& \multicolumn{4}{c|}{NoXi}
& \multicolumn{4}{c|}{UDIVA}
& \multicolumn{4}{c}{SeamInt} \\ \cmidrule{2-13}
& Face & Body & Audio & Text
& Face & Body & Audio & Text
& Face & Body & Audio & Text \\
\midrule
(a) L
& 39.7\std{8.9} & 39.9\std{8.8} & 42.2\std{6.0} & 41.4\std{1.4}
& 44.4\std{4.0} & 45.4\std{4.7} & 47.4\std{5.1} & 49.1\std{3.1}
& 32.8\std{0.3} & 43.2\std{2.7} & 37.5\std{1.9} & 39.3\std{9.5} \\

(b) H
& 45.8\std{5.8} & 47.7\std{4.7} & 47.8\std{2.6} & 47.3\std{5.3}
& 52.2\std{1.7}* & 50.8\std{1.0} & 53.4\std{3.2}* & 52.8\std{1.9}
& 49.8\std{3.9}* & 50.1\std{3.3}* & 49.1\std{9.4} & 47.6\std{5.2} \\

(c) LH+MoI
& 51.3\std{4.7}* & 53.5\std{2.1} & 52.7\std{1.4} & 51.8\std{5.9}*
& 64.3\std{8.4} & 61.6\std{4.8} & 64.7\std{5.9}* & 64.5\std{7.2}*
& 54.5\std{7.5} & 53.7\std{3.3} & 56.5\std{4.3} & 55.9\std{8.5} \\

(d) LH+MoI+HC
& \textbf{61.5}\std{2.6}* & \textbf{60.7}\std{3.5}* & \textbf{61.3}\std{5.6} & \textbf{61.0}\std{5.2}
& \textbf{64.6}\std{3.5}* & \textbf{62.3}\std{0.8} & \textbf{65.6}\std{5.3} & \textbf{65.1}\std{0.4}
& \textbf{63.8}\std{1.7} & \textbf{65.7}\std{4.5} & \textbf{64.5}\std{5.7} & \textbf{65.9}\std{6.7} \\
\bottomrule
\end{tabular}

\begin{tablenotes}[flushleft]
\scriptsize
\item[] $\ast$ indicates a statistically significant improvement over the preceding configuration ($p<0.05$).
\end{tablenotes}
\end{threeparttable}
\vspace{-1.5em}
\end{table*}

Table~\ref{Table: 3} shows progressive improvements as the relational representation becomes increasingly structured. High-level features consistently outperform low-level features, indicating that semantic behavioural cues provide stronger evidence for relationship recognition. Combining low- and high-level features with MoI-Former further improves the performance from 49.0\% to 55.1\% on NoXi, 55.9\% to 65.9\% on UDIVA, and 55.3\% to 62.4\% on SeamInt, demonstrating the benefit of structured interaction reasoning over individual feature levels. Adding HC-Loss achieves the best results on all benchmarks, with larger gains on NoXi and SeamInt where relationships exhibit richer semantic hierarchies. In contrast, the improvement on UDIVA is limited because its binary Known/Unknown formulation does not involve an explicit relationship hierarchy.

Table~\ref{Table: 4} further evaluates different modalities. The consistent improvements from \textbf{L} to \textbf{H}, \textbf{LH+MoI}, and \textbf{LH+MoI+HC} across modalities demonstrate the complementary roles of semantic feature levels, interaction reasoning, and hierarchical supervision. For example, the NoXi face branch improves from $39.7$ to $45.8$, $51.3$, and $61.5$, while the SeamInt audio branch increases from $37.5$ to $49.1$, $56.5$, and $64.5$. On UDIVA, the gains from HC-Loss are relatively smaller, consistent with its binary Known/Unknown setting where the relationship hierarchy is not explicitly defined. Overall, these results support the complementary roles of semantic feature hierarchy, cross-participant interaction reasoning, and hierarchical supervision in HiRe-DRR.

\begin{figure}[ht]
    \vspace{-0.5em}
    \setlength{\abovecaptionskip}{0.cm}
    \setlength{\belowcaptionskip}{-0.cm}
    \centering
    \includegraphics[width=2.5in]{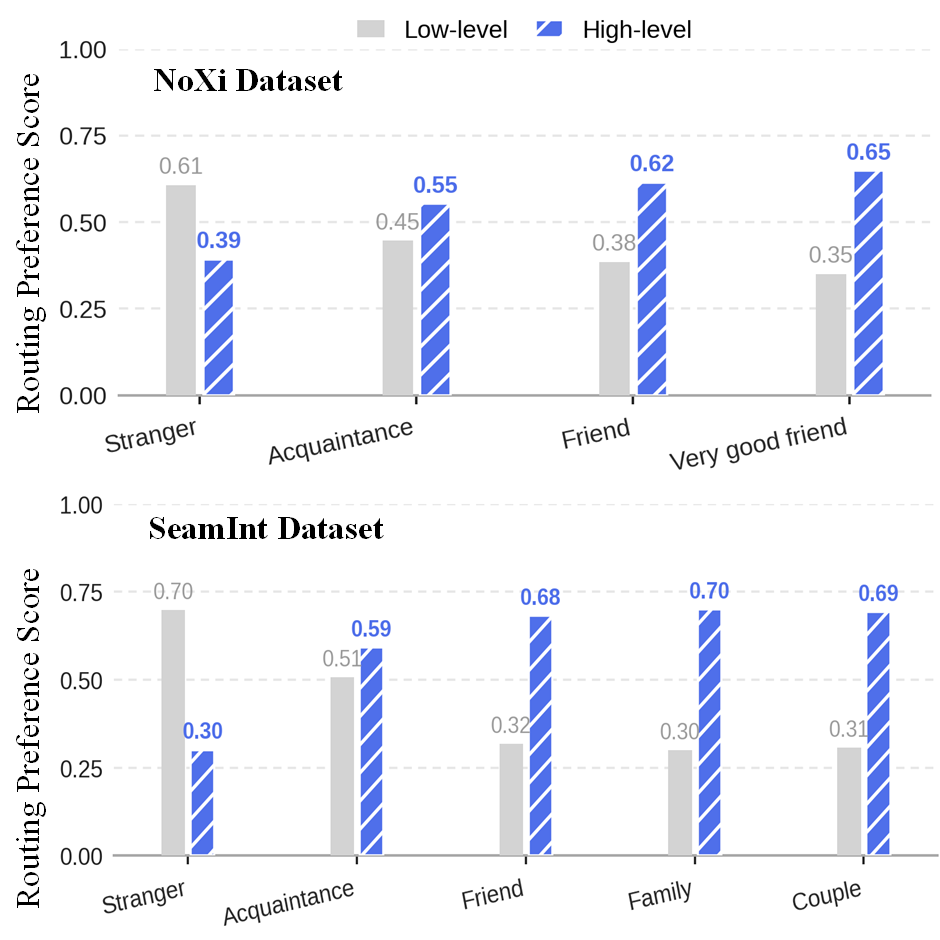}
    \caption{Class-wise routing preferences over low- and high-level interactions on NoXi and SeamInt datasets.
    \label{Fig: 6}}
    \vspace{-0.5em}
\end{figure}
\begin{figure}[ht]
    \vspace{-0.5em}
    \setlength{\abovecaptionskip}{0.cm}
    \setlength{\belowcaptionskip}{-0.cm}
    \centering
    \includegraphics[width=2.5in]{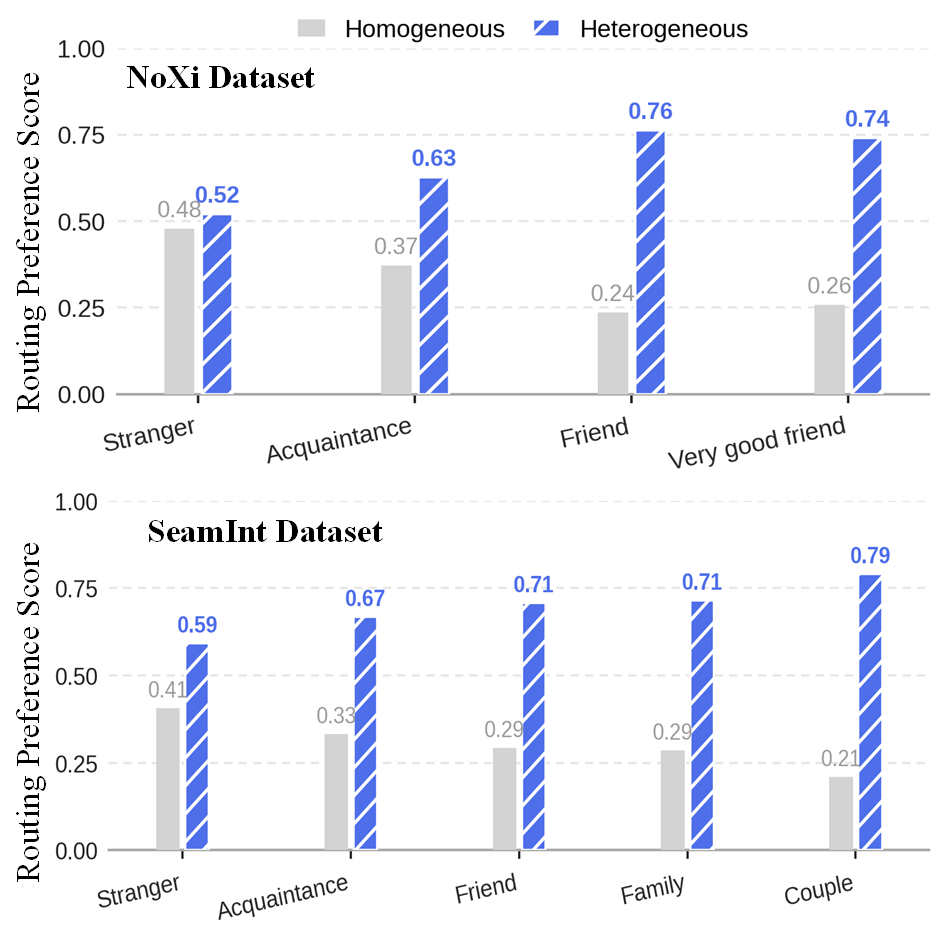}
    \caption{Class-wise routing preferences over homogeneous and heterogeneous interactions on NoXi and SeamInt datasets.
    \label{Fig: 7}}
    \vspace{-1.0em}
\end{figure}

\subsubsection{Ablation of the Feature Level}
We analyse how MoI-Former distributes routing preferences between low- and high-level interactions across relationship categories. For visualisation, the routing preference is obtained by averaging the router probabilities across all MoI encoder layers. All routing preferences reported in the following analyses are computed using the same procedure. As shown in Fig.~\ref{Fig: 6}, both datasets exhibit a consistent shift towards high-level evidence as relationships become socially closer. This trend reflects the complementary roles of different feature levels: low-level cues capture observable behavioural correspondences for coarse social distinctions, whereas high-level cues provide affective and semantic evidence for fine-grained relationship understanding. The persistent contribution of both levels indicates that MoI-Former adaptively adjusts evidence granularity according to relationship semantics rather than relying on fixed multimodal fusion.

\subsubsection{Ablation of the Interaction Type}
We further investigate how MoI-Former distributes routing preferences between homogeneous and heterogeneous interactions across relationship categories. Homogeneous interactions capture dependencies between the same modalities across participants (e.g., face-face correspondence), whereas heterogeneous interactions model cross-modal responses between participants (e.g., one participant's speech associated with the other's facial or bodily reactions). Fig.~\ref{Fig: 7}, both datasets exhibit a consistent shift towards heterogeneous interactions for socially closer relationships. This pattern suggests that coarse relationship distinctions can be supported by direct behavioural consistency within the same modality, whereas fine-grained relationship understanding requires cross-modal evidence that captures complementary expressions and responses between participants. The remaining homogeneous preference across all categories further indicates that MoI-Former integrates both interaction types adaptively rather than relying exclusively on heterogeneous reasoning.

\begin{figure*}[t!]
    \vspace{-0.5em}
    \setlength{\abovecaptionskip}{0.cm}
    \setlength{\belowcaptionskip}{-0.cm}
    \centering
    \includegraphics[width=\textwidth]{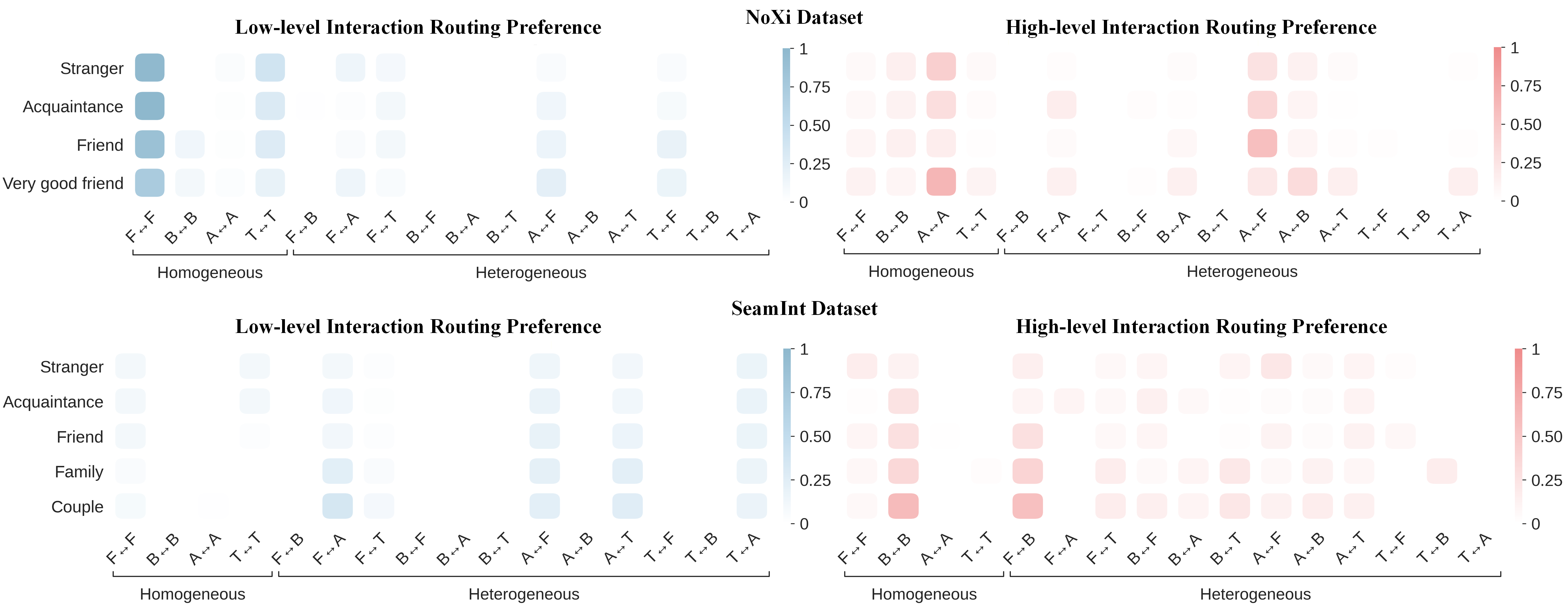}
    \caption{Class-wise routing preferences of MoI-Former over low- and high-level homogeneous and heterogeneous interactions. F, B, A, and T denote face, body, audio, and text, respectively; darker cells indicate larger mean routing scores.
    \label{Fig: 8}}
    \vspace{-1.0em}
\end{figure*}

\begin{table}[t!]
\vspace{-0.5em}
\centering
\footnotesize
\caption{Token masking based on routing-preference.}
\vspace{-0.5em}
\label{Table: 5}
\setlength{\tabcolsep}{0.05cm}
\renewcommand{\arraystretch}{1.20}
\begin{threeparttable}
\begin{tabular}{lccc}
\toprule
Configuration & NoXi & UDIVA & SeamInt \\
\midrule
Full model 
& \textbf{63.1}\std{5.6}
& \textbf{66.8}\std{6.4}
& \textbf{68.5}\std{2.6} \\

Mask high-preference tokens 
& 61.7\std{4.6} (-1.4)
& 65.0\std{6.0} (-1.8)
& 67.0\std{5.4} (-1.5) \\

Mask low-preference tokens 
& 63.8\std{5.0} (+0.7)
& 67.0\std{3.3} (+0.2)
& 68.8\std{3.3} (+0.3) \\
\bottomrule
\end{tabular}

\begin{tablenotes}[flushleft]
\footnotesize
\item \textit{Note:} The results are reported in mAP (\%). Values in parentheses show the percentage change relative to the full model.
\end{tablenotes}
\end{threeparttable}
\vspace{-1em}
\end{table}

\begin{figure}[t!]
    \vspace{-0.5em}
    \setlength{\abovecaptionskip}{0.cm}
    \setlength{\belowcaptionskip}{-0.cm}
    \centering
    \includegraphics[width=3.0in]{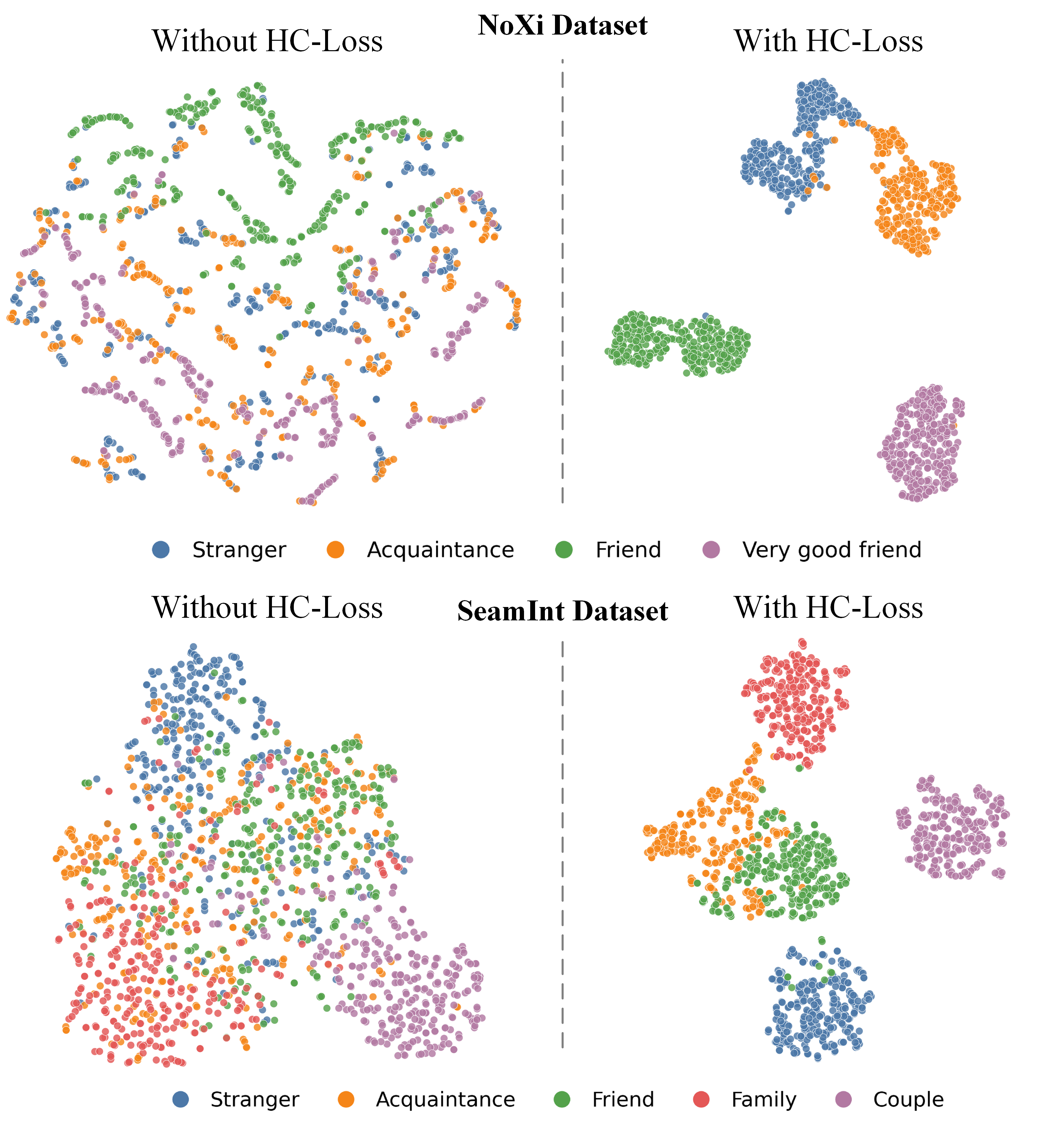}
    \vspace{-0.5em}
    \caption{t-SNE projections of relationship representations learned with and without HC-Loss on NoXi and SeamInt datasets.
    \label{Fig: 9}}
    \vspace{-1em}
\end{figure}

In addition, we analyse how homogeneous and heterogeneous interactions contribute to different relationships. Figure~\ref{Fig: 8} shows that MoI-Former dynamically adjusts interaction preferences according to feature levels and relationship categories rather than relying on a fixed structure.

On NoXi, low-level routing consistently favours homogeneous F$\leftrightarrow$F and T$\leftrightarrow$T interactions, indicating that direct facial correspondence and conversational alignment provide fundamental cues for relationship recognition. High-level routing is more category-dependent: distant relationships, such as \textit{Stranger} and \textit{Acquaintance}, rely more on homogeneous A$\leftrightarrow$A interactions, whereas \textit{Very Good Friend} increasingly activates heterogeneous A$\leftrightarrow$F and A$\leftrightarrow$B interactions. This suggests that closer relationships require complementary affective and behavioural responses beyond single-modality consistency.

SeamInt exhibits stronger cross-modal interaction patterns, where low-level routing already involves heterogeneous pathways such as F$\leftrightarrow$A and A$\leftrightarrow$T. At the high level, intimate relationships further activate diverse interactions, including B$\leftrightarrow$B, F$\leftrightarrow$B, and B$\leftrightarrow$T, suggesting that \textit{Family} and \textit{Couple} rely on coordinated behavioural, affective, and semantic responses. Despite dataset variations, a consistent trend emerges: distant relationships depend more on direct interaction cues, whereas closer relationships involve richer cross-modal configurations. These findings demonstrate the importance of relationship-aware interaction selection and reveal distinct interaction patterns underlying different relationship categories.

Table~\ref{Table: 5} further validates the routing patterns through recognition performance difference. Masking the five high-preference tokens consistently reduces mAP, whereas masking the five low-preference tokens causes no degradation and even leads to slight improvements across all datasets. These results indicate that highly routed interactions provide stronger recognition evidence, while low-preference tokens may contain redundant information.

\subsubsection{Ablation of the HC-Loss}
To analyse the effect of HC-Loss on relationship representation learning, we visualise the latent features before the final classifier using t-SNE~\cite{76}, as shown in Fig.~\ref{Fig: 9}. Without HC-Loss, the learned representations exhibit substantial inter-class overlap and dispersed distributions, indicating that flat classification supervision provides limited guidance for capturing the hierarchical structure of relationship categories.

With HC-Loss, in contrast, the feature space becomes more compact and discriminative, with clearer boundaries between relationship classes. HC-Loss enhances the discrimination of semantically adjacent classes by encouraging representations that better reflect the underlying relationship hierarchy.

\subsection{Qualitative Analyses}
\begin{figure*}[t!]
    \vspace{-0.5em}
    \setlength{\abovecaptionskip}{0.cm}
    \setlength{\belowcaptionskip}{-0.cm}
    \centering
    \includegraphics[width=\textwidth]{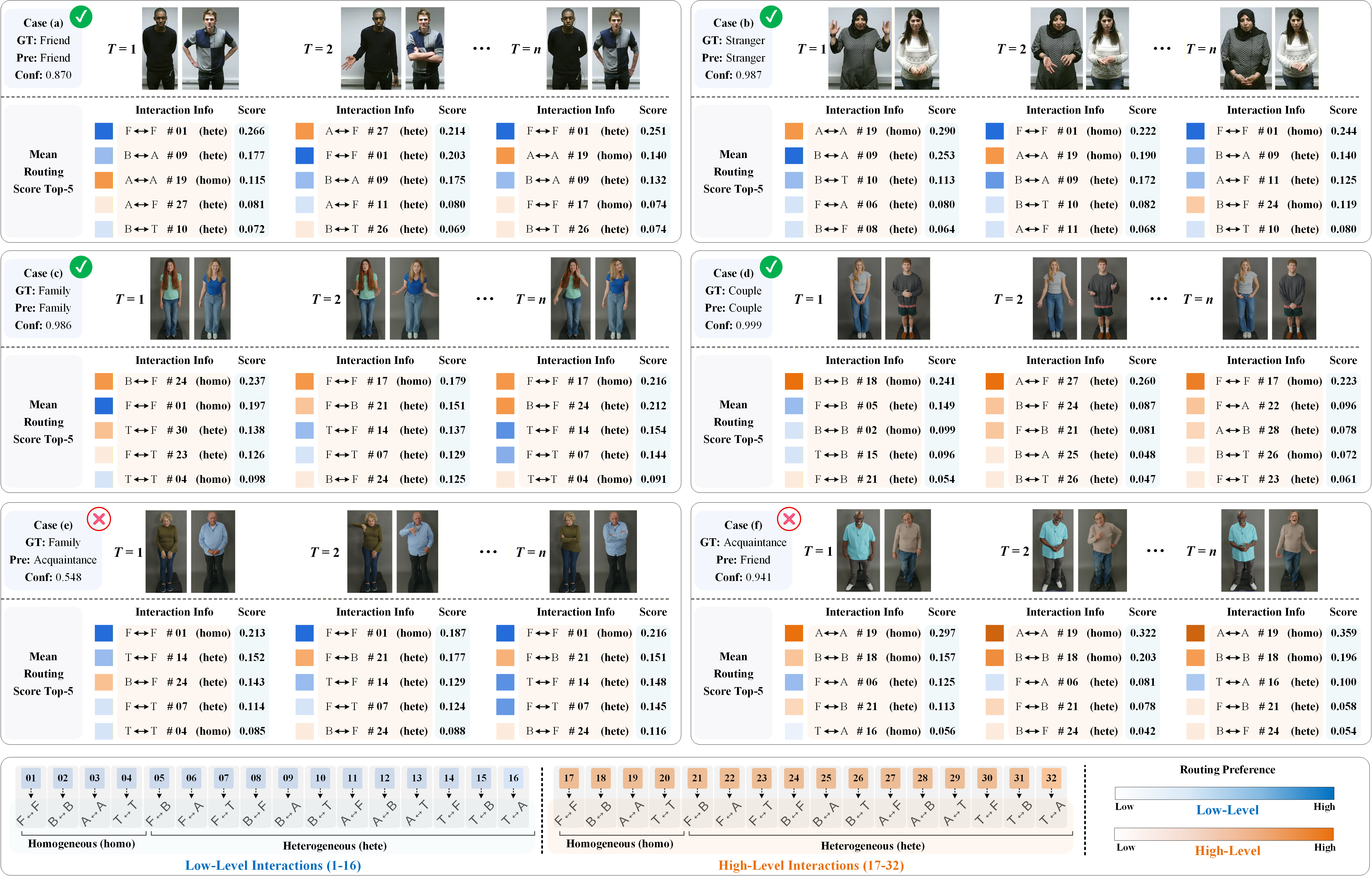}
    \vspace{-1.0em}
    \caption{Qualitative analysis of MoI-Former routing on NoXi and Seamless Interaction. Cases (a)-(b) are from NoXi, and Cases (c)-(f) are from Seamless Interaction. The routed Top-$5$ interactions reveal relationship-dependent evidence selection, where blue and orange indicate low- and high-level interactions and colour intensity reflects routing preference. Correct and failure cases illustrate the effectiveness and limitations of sparse interaction reasoning under ambiguous relationship patterns. \textit{Note: GT: Ground Truth, Pre: Prediction, Conf: Confidence}.  \label{Fig: 10}}
    \vspace{-1.0em}
\end{figure*}

We qualitatively analyse six dyadic interaction examples in Fig.~\ref{Fig: 10}. Cases (a)-(d) are correctly recognised, while Cases (e) and (f) illustrate failure cases. Instead of uniformly aggregating multimodal cues, MoI-Former sparsely routes interaction tokens and reveals how evidence with different semantic abstraction levels contributes to relationship recognition.

In the successful cases, the routed interactions align with the semantic requirements of the target relationship. For Case (c), the recognition of \emph{Family} relies more on high-level behavioural and affective cues, where heterogeneous pathways capture complementary responses between participants. Similarly, Case (d) activates semantically structured facial, bodily, and cross-modal interactions to characterise \emph{Couple} beyond observable motion patterns. In contrast, Case (b) mainly relies on low-level interactions that capture directly observable behavioural correspondences associated with the larger social distance of \emph{Stranger}. These examples demonstrate that MoI-Former adaptively selects evidence at different abstraction levels according to the relational distinction, rather than consistently favouring specific modalities.

The failure cases further reveal the challenge caused by overlapping interaction patterns among different relationship categories. In Case (e), although the ground-truth relationship is \emph{Family}, the routing preference is dominated by low-level F$\leftrightarrow$F, T$\leftrightarrow$F, and F$\leftrightarrow$B pathways, which capture simple facial engagement, conversational exchange, and behavioural coordination. Such cues are common indicators of familiarity and social interaction, but they may also appear in less intimate relationships, providing insufficient evidence to distinguish \emph{Family} from \emph{Acquaintance}. In contrast, Case (f) exhibits stronger high-level interaction patterns, including A$\leftrightarrow$A, B$\leftrightarrow$B, and heterogeneous body-related pathways. These interactions reflect richer emotional and behavioural coordination, which are often associated with closer relationships. However, similar high-level responses may occasionally occur in acquaintance interactions, causing the model to overestimate relational closeness and confuse \emph{Acquaintance} with \emph{Friend}. These cases indicate that relationship recognition remains challenging when discriminative relational cues are entangled with interaction patterns shared across categories.

\section{Discussion and Conclusion}
\textbf{Discussion.} HiRe-DRR reveals that dyadic relationships can be understood through relationship-dependent interaction evidence rather than uniformly aggregated multimodal cues. The routing analysis indicates that different relationships rely on distinct interaction patterns, suggesting that future socially intelligent systems may benefit from explicitly modelling the behavioural and semantic cues underlying interpersonal relationships. In human-robot interaction, such relationship-aware understanding can provide additional context for adapting communication strategies and interaction behaviours to different social partners~\cite{01,02}. In social assistance and interactive agents, recognising relationship-specific interaction patterns can support more personalised and context-aware responses~\cite{05}. Beyond these scenarios, the discovered interaction patterns may also provide useful relational context for broader applications, including human activity understanding~\cite{82} and social interaction analysis~\cite{10}.

\textbf{Conclusion.} We introduced HiRe-DRR, a hierarchical multimodal interaction reasoning framework for dyadic relationship recognition. By organising relational evidence across feature level, cross-participant interaction type, and relationship hierarchy, HiRe-DRR preserves relationship-specific cues that may be obscured by conventional multimodal fusion. MoI-Former adaptively identifies relationship-relevant interactions, while HC-Loss exploits the hierarchical nature of social relationships through coarse-to-fine supervision. Experiments on three benchmarks demonstrate consistent improvements over representative image-, video-, and multimodal methods. Routing analyses further reveal that distant relationships rely more on observable behavioural correspondences, whereas closer relationships increasingly involve high-level semantic and heterogeneous cross-modal interactions.

Despite these advances, recognising dyadic relationships in real-world scenarios remains challenging. Existing benchmarks mainly capture short-term interactions, while human relationships evolve through long-term behavioural patterns, changing contexts, and broader social structures. Moreover, socially adjacent relationships may share similar interaction evidence due to individual communication styles and situational variations. Future work will explore long-range temporal relationship modelling, multi-person relational reasoning, and more adaptive evidence discovery mechanisms to support robust social understanding in complex environments.

\bibliographystyle{IEEEtran}
\bibliography{Bibliography}

\vspace{1.5em}

\textbf{Wang Tang} received the Ph.D. degree from the Sichuan University, China. He is currently an Associate Professor with the Chengdu University of Information Technology. He was a visiting scholar at the Cambridge AFAR Lab, University of Cambridge. His research interests include image processing, computer vision, and pattern recognition.
\vspace{0.5em}

\textbf{Fethiye Irmak Dogan} is a Postdoctoral Researcher in the AFAR Lab at the University of Cambridge (UK). She received her Ph.D. degree in Computer Science from KTH Royal Institute of Technology (Sweden). Her research focuses on human-robot interaction, robot learning, and explainability.
\vspace{0.1em}

\textbf{Linbo Qing} received the Ph.D. degree from Sichuan University, China. He is currently a Professor with the College of Electronics and Information Engineering, Sichuan University. His research interests include artificial intelligence, computer vision, image processing, and digital health.
\vspace{0.5em}

\textbf{Ce Zhu (Fellow, IEEE)} received the Ph.D. degree from Southeast University, China. He is currently a Professor with the University of Electronic Science and Technology of China. His research interests include video coding, video analysis and processing, visual perception, and multimedia applications.

\vspace{0.5em}
\textbf{Hatice Gunes (Senior Member, IEEE)} is a Full Professor of Affective Intelligence and Robotics (AFAR) in the Department of Computer Science and Technology, University of Cambridge, leading the \href{https://cambridge-afar.github.io/} {Cambridge AFAR Lab}. She is a former President of the Association for the Advancement of Affective Computing, a former Faculty Fellow of the Alan Turing Institute, and is currently a Fellow of the EPSRC and a Staff Fellow of Trinity Hall.

\end{document}